%% file: neurips_2025_GDM_format.tex
\documentclass[11pt, a4paper, logo, copyright]{googledeepmind}

\usepackage[utf8]{inputenc} 
\usepackage[T1]{fontenc}    
\usepackage{url}            
\usepackage{booktabs}       
\usepackage{amsfonts}       
\usepackage{nicefrac}       
\usepackage{microtype}      
\usepackage{xcolor}         
\usepackage{commands}
\usepackage{amsmath}
\usepackage{algorithm}
\usepackage{algpseudocode}
\usepackage{url}
\usepackage{xr}
\usepackage{listings}
\usepackage{adjustbox}
\usepackage{subcaption}
\usepackage{xspace}
\usepackage{longtable}
\usepackage{caption}
\usepackage{xcolor}
\usepackage{tcolorbox}
\usepackage{xskak}
\usepackage{multicol}
\usepackage{multirow}
\usepackage{wrapfig}

\usepackage{cleveref}
\usepackage[authoryear, sort&compress, round]{natbib}
\usepackage{csquotes}

\renewcommand{\today}{October 27, 2025}

\newcommand{\amatzia}{\textbf{AA} }
\newcommand{\jonathan}{\textbf{JL} }
\newcommand{\matthew}{\textbf{MS} }

\newcommand{\appendixchessboardwidth}{1.0}
\newcommand{\mainpuzzlewidth}{0.8}

\title{Evaluating In Silico Creativity:\\An Expert Review of AI Chess Compositions}

\author[1]{Vivek Veeriah}
\author[2, $\dagger$]{Federico Barbero}
\author[1]{Marcus Chiam}
\author[1]{Xidong Feng}
\author[1]{Michael Dennis}
\author[3]{Ryan Pachauri}
\author[1]{Thomas Tumiel}
\author[4, $\dagger$]{Johan Obando-Ceron}
\author[1]{Jiaxin Shi}
\author[1]{Shaobo Hou}
\author[1]{Satinder Singh}
\author[1]{Nenad Toma\v{s}ev}
\author[1]{Tom Zahavy}

\affil[1]{Google DeepMind}
\affil[2]{University of Oxford}
\affil[3]{Google}
\affil[4]{Mila, University of Montreal}
\affil[$\dagger$]{During a Google DeepMind internship.}

\begin{abstract}\end{abstract}

\begin{document}

\maketitle


\vspace{-1.5cm}

The rapid advancement of Generative AI has raised significant questions regarding its ability to produce creative and novel outputs. Our recent work investigates this question within the domain of chess puzzles and presents an AI system designed to generate puzzles characterized by aesthetic appeal, novelty, counter-intuitive and unique solutions. We briefly discuss our method below and refer the reader to the \href{https://drive.google.com/drive/folders/1bdj34gBUoWkqEC_CDW64HuIIIKWSKBNX?usp=drive_link}{technical paper} for more details. To assess our system's creativity, we presented a curated booklet of AI-generated puzzles to three world-renowned experts: International Master for chess compositions 
\href{https://players.chessbase.com/en/player/Avni_Amatzia/13285}{Amatzia Avni}, Grandmaster \href{http://www.jlevitt.dircon.co.uk/index.htm}{Jonathan Levitt}, and Grandmaster \href{https://matthewsadler.me.uk/}{Matthew Sadler}. All three are noted authors on chess aesthetics and the evolving role of computers in the game. They were asked to select their favorites and explain what made them appealing, considering qualities such as their creativity, level of challenge, or aesthetic design.
This paper presents these selected puzzles, integrating the experts' analysis to explore what makes them counter-intuitive and beautiful.

\noindent \textbf{Generating Millions of Chess Puzzles:} Our method involves training generative neural networks (Auto-Regressive Transformer, Discrete Diffusion, and MaskGit) on a dataset of 4 million chess puzzles from Lichess\footnote{https://database.lichess.org/\#puzzles} to learn the distribution of those puzzles. Each position was encoded as a sequence using Forsyth-Edwards Notation (FEN), and a neural network was trained to predict the distribution of the next character in the string based on the characters that preceded it. The trained network was then employed as a generative model to sample chess puzzles, starting from the first character of the FEN and iteratively sampling the remaining characters.  We further trained the generative neural network with reinforcement learning. This involved a custom reward design, which was used to select the best samples and iteratively train the network to generate puzzles with higher rewards. The reward function had two parts: a uniqueness check, similar to the one used in Lichess, to ensure there was only one winning move; and a counter-intuitiveness check, to ensure the position could be solved by a strong chess engine but not a weak one. 

\noindent \textbf{Selecting Puzzles by Reward:} We generated approximately 4 million chess positions from the aforementioned models and filtered them using a hybrid approach. Positions were first ranked by a reward function, then processed by aesthetic theme detectors. While the detectors were imprecise alone, their effectiveness was greatly enhanced by the initial reward-based ranking. To find good puzzles, we manually reviewed the top 50 samples for each theme, a process validated with FIDE players in the 2200 - 2300 rating range.

\noindent \textbf{Chess Booklet:} A collection of selected puzzles was compiled into a \hyperref[sec:booklet]{booklet} and was sent for review. The experts' feedback was generally positive; they noted the innovative fusion of aesthetic themes and the "over-the-board" vision. Nevertheless, the reviewers also provided constructive criticism, pointing out that some positions were trivial, while the collection overall lacked the profundity and complexity of traditional endgame studies. They also remarked that certain puzzles were unrealistic. For future development, they recommended increasing the complexity and depth of the positions, incorporating problems with more complex sidelines and robust counter-play, and wanted to see more surprising theme combinations.

\noindent \textbf{Creativity in Chess:} For the scope of this work, we broadly identify a chess puzzle as creative when its solution contains a sense of surprise, challenge, and beauty. Of course, a precise measure of creativity is hard to define and largely subjective -- we found in our study that even very strong chess experts frequently differ in their assessments of a puzzle's creative merit. There are many factors that may influence the assessment, for instance, the player's skill level, prior exposure to similar tactical patterns, and general individual aesthetic preferences. A puzzle may be deemed creative if it presents an uncommon exposition of a familiar theme (e.g. Puzzle \ref{fig:puzzle_8} contains the classic smothered mate theme) or if its solution requires counter-intuitive moves (e.g. Puzzle \ref{fig:main_puzzle}). 

In the next section, we present a series of puzzles that were highlighted by the experts. The solutions and expert commentary for these puzzles are provided in the subsequent sections, giving the reader an opportunity to attempt them first.

\noindent{\large\textbf{Highlighted Puzzles without Solutions}}


\begin{figure*}[h]
\centering
    \begin{subfigure}{0.32\textwidth}
    \centering
    \adjustbox{max width=\mainpuzzlewidth\textwidth}
        {
            \newchessgame[setfen=1r1r2k1/Q2p1R1p/2p2R2/1p3pB1/1P4q1/8/5K2/8 w]
            \setchessboard{smallboard,color=yellow!70,showmover=true,mover=w}
            \chessboard
        }
        \caption{
            Puzzle 1: \textbf{\href{https://lichess.org/analysis/1r1r2k1/Q2p1R1p/2p2R2/1p3pB1/1P4q1/8/5K2/8 w}{[Analyse on Lichess]}
        }}
        \end{subfigure}
    ~
    \begin{subfigure}{0.32\textwidth}
    \centering
    \adjustbox{max width=\mainpuzzlewidth\textwidth}
        {
            \newchessgame[setfen=1qb5/5k2/P1Qp1bNp/2pP1P2/2P1pP1P/8/3rB1R1/4K3 b]
            \setchessboard{smallboard,color=yellow!70,showmover=true,mover=b}
            \chessboard
        }
        \caption{
            Puzzle 2: \textbf{\href{https://lichess.org/analysis/1qb5/5k2/P1Qp1bNp/2pP1P2/2P1pP1P/8/3rB1R1/4K3 b}{[Analyse on Lichess]}
        }}
        \end{subfigure}
    ~
    \begin{subfigure}{0.32\textwidth}
    \centering
    \adjustbox{max width=\mainpuzzlewidth\textwidth}
        {
            \newchessgame[setfen=rnbqrbk1/pp3Rp1/2p1p1N1/3p1P1Q/3PnB2/2P5/PP3P1P/6K1 w]
            \setchessboard{smallboard,color=yellow!70,showmover=true,mover=w}
            \chessboard
        }
        \caption{
            Puzzle 3: \textbf{\href{https://lichess.org/analysis/rnbqrbk1/pp3Rp1/2p1p1N1/3p1P1Q/3PnB2/2P5/PP3P1P/6K1 w}{[Analyse on Lichess]}
        }}
        \end{subfigure}
\end{figure*}

\begin{figure*}[h]
\centering
    \begin{subfigure}{0.32\textwidth}
    \centering
    \adjustbox{max width=\mainpuzzlewidth\textwidth}
        {
            \newchessgame[setfen=r4b1k/pq1PN1pp/nn1Q4/2p3P1/2p4P/1Pp5/P7/2KR4 w]
            \setchessboard{smallboard,color=yellow!70,showmover=true,mover=w}
            \chessboard
        }
        \caption{
            Puzzle 4: \textbf{\href{https://lichess.org/analysis/r4b1k/pq1PN1pp/nn1Q4/2p3P1/2p4P/1Pp5/P7/2KR4 w}{[Analyse on Lichess]}
        }}
        \end{subfigure}
        ~
        \begin{subfigure}{0.32\textwidth}
    \centering
    \adjustbox{max width=\mainpuzzlewidth\textwidth}
        {
            \newchessgame[setfen=8/4p3/p4p2/P1K2b2/BP1p2k1/2P1n3/3P2P1/8 w]
            \setchessboard{smallboard,color=yellow!70,showmover=true,mover=w}
            \chessboard
        }
        \caption{
            Puzzle 5: \textbf{\href{https://lichess.org/analysis/8/4p3/p4p2/P1K2b2/BP1p2k1/2P1n3/3P2P1/8 w}{[Analyse on Lichess]}
        }}
        \end{subfigure}
    ~
    \begin{subfigure}{0.32\textwidth}
    \centering
    \adjustbox{max width=\mainpuzzlewidth\textwidth}
        {
            \newchessgame[setfen=8/3k1p2/3Pb3/2K5/7p/p7/3R2P1/8 b]
            \setchessboard{smallboard,color=yellow!70,showmover=true,mover=b}
            \chessboard
        }
        \caption{
            Puzzle 6: \textbf{\href{https://lichess.org/analysis/8/3k1p2/3Pb3/2K5/7p/p7/3R2P1/8 b}{[Analyse on Lichess]}
        }}
        \end{subfigure}
\end{figure*}

\begin{figure*}[h]
\centering
    \begin{subfigure}{0.32\textwidth}
    \centering
    \adjustbox{max width=\mainpuzzlewidth\textwidth}
        {
            \newchessgame[setfen=1q4rk/ppr1PQpp/1b3R2/3R4/1P6/4P3/P5PP/6K1 w]
            \setchessboard{smallboard,color=yellow!70,showmover=true,mover=w}
            \chessboard
        }
        \caption{
            Puzzle 7: \textbf{\href{https://lichess.org/analysis/1q4rk/ppr1PQpp/1b3R2/3R4/1P6/4P3/P5PP/6K1 w}{[Analyse on Lichess]}
        }}
        \end{subfigure}
    ~
    \begin{subfigure}{0.32\textwidth}
    \centering
    \adjustbox{max width=\mainpuzzlewidth\textwidth}
        {
            \newchessgame[setfen=6rk/Q7/3q4/5p2/2PP1P2/P5Pr/7P/R4RK1 b]
            \setchessboard{smallboard,color=yellow!70,showmover=true,mover=b}
            \chessboard
        }
        \caption{
            Puzzle 8: \textbf{\href{https://lichess.org/analysis/6rk/Q7/3q4/5p2/2PP1P2/P5Pr/7P/R4RK1 b}{[Analyse on Lichess]}
        }}
        \end{subfigure}
    ~
    \begin{subfigure}{0.32\textwidth}
    \centering
    \adjustbox{max width=\mainpuzzlewidth\textwidth}
        {
            \newchessgame[setfen=r4r1k/6pp/4Q3/5pN1/p2P1P2/1P5P/q5P1/2R4K w]
            \setchessboard{smallboard,color=yellow!70,showmover=true,mover=w}
            \chessboard
        }
        \caption{
            Puzzle 9: \textbf{\href{https://lichess.org/analysis/r4r1k/6pp/4Q3/5pN1/p2P1P2/1P5P/q5P1/2R4K w}{[Analyse on Lichess]}
        }}
        \end{subfigure}
\end{figure*}

\setcounter{figure}{0}

\clearpage
\noindent{\large\textbf{Comments from Chess Experts}}

The experts' diverse preferences for the puzzles underscore the highly subjective nature of beauty and creativity in chess, as they rarely agreed on which were the most compelling. This review will begin with overall commentary on the booklet before discussing the one puzzle that earned unanimous praise. From there, we will delve into the specific puzzles that appealed to each expert individually, along with their comments.

\begin{quote}
\noindent\textbf{IM for Chess Compositions Amatzia Avni:} This booklet adds novel, AI-generated puzzles to the existing chess literature, serving as a resource for both training and enjoyment. A valuable chess puzzle should be original and creative, with a surprising, counter-intuitive key move and a smart follow-up. The ideal puzzle is also aesthetically pleasing and offers a satisfying, flowing solution. It must strike a good balance in difficulty -- challenging enough to avoid being obvious, but not so hard as to cause frustration. I selected Puzzles \ref{fig:puzzle_1}, \ref{fig:puzzle_2}, \ref{fig:puzzle_3} as they met my expectations and have described them in the next section.
\end{quote}

\begin{quote}
\noindent\textbf{GM Jonathan Levitt:} For years, chess composers have worked with computers to verify soundness of their work and thus assist in the process of creation too. This nature of collaboration is evolving, with AI now capable of generating interesting chess positions, beyond just ``mining'' databases. The positions in this booklet represent a pioneering step in this human-AI partnership. While these initial AI-generated endgame compositions are not yet at a prize-winning level, they clearly demonstrate the potential to be. I have highlighted some of my selections as Puzzles \ref{fig:puzzle_4} and \ref{fig:puzzle_5} in the next section.
\end{quote}

\begin{quote}
\noindent\textbf{GM Matthew Sadler:} It was an intriguing experience to assess the chess booklet. I have strong preferences about what makes a good puzzle position. In particular, I favor natural positions resulting from reasonable play by both sides. Puzzles lose my interest if one side's pieces are clearly misplaced or if a complex solution yields a minimal advantage, like being up a single pawn after sacrificing multiple pieces. Even with those stringent conditions, I enjoyed many of the positions in this booklet of which I will highlight a small selection (Puzzles \ref{fig:puzzle_6}, \ref{fig:puzzle_7}, \ref{fig:puzzle_8}) in the next section.
\end{quote}

We will be referring to IM for chess compositions Amatzia Avni as \textbf{AA}, GM Jonathan Levitt as \textbf{JL} and GM Matthew Sadler as \textbf{MS}.

\begin{wrapfigure}[13]{r}{0.3\textwidth}
\setchessboard{boardfontsize=12pt,labelfontsize=6pt,showmover=true,}
\newchessgame[setfen=1r1r2k1/Q2p1R1p/2p2R2/1p3pB1/1P4q1/8/5K2/8 w]
    \chessboard    
\caption{
AI-generated puzzle with unanimous acclaim from the experts.
\textbf{\href{https://lichess.org/analysis/1r1r2k1/Q2p1R1p/2p2R2/1p3pB1/1P4q1/8/5K2/8 w}{[Analyse on Lichess]}}
}
\end{wrapfigure}


\noindent {\large\textbf{A Puzzle That Intrigued All the Experts:}}

The puzzle position that all experts agreed to be beautiful is shown on the right. White has aggressively placed pieces, but an exposed King on f2 and a misplaced Queen on a7. White has to manage to mount an attack that does not allow counter-play. Only one move in the position achieves all this. 

The winning move is \mainline{1.Rg6+!}, which \textbf{AA}, \jonathan and \matthew described as ``unorthodox'' and by ``no means natural or obvious sacrifice''. The move starts the attack by giving up both rooks at once! \amatzia further described the move as ``certainly not what you would take into considering as the first candidate move''.

After \variation{1. Rg6+!}, either rook can be captured: After \variation{1...Kxf7} or \variation{1... hxg6}, the continuations are similar. Choosing to capture with the King \variation{1...Kxf7} leads to \variation{2. Qa1!} and after black captures the second rook \variation{2...hxg6 3. Qf6+}, white has an unstoppable attack and is covering all of the checks from the black Queen. The puzzle is paradoxical as the solution involves sacrificing \emph{both} very active Rooks to prepare the slow repositioning and infiltration of the misplaced Queen on a7. Further, the combinations are geometric and involve participation of pieces from both flanks. They are difficult to find even for a strong player.

\setcounter{figure}{0}    

\begin{figure}[h]
\centering
    \begin{subfigure}[t]{0.48\textwidth}
    \centering
    \adjustbox{max width=\textwidth}
        {
            \newchessgame[setfen=1r1r2k1/Q2p1R1p/2p3R1/1p3pB1/1P4q1/8/5K2/8 b - - 1 1]
            \setchessboard{boardfontsize=12pt,labelfontsize=6pt,color=yellow!70,showmover=true,mover=b} 
            \chessboard
        }
        \end{subfigure}
    ~
    \begin{subfigure}[t]{0.48\textwidth}
    \centering
    \adjustbox{max width=\textwidth}
        {
            \newchessgame[setfen= 1r1r2k1/3p4/2p2QpB/1p3p2/1P4q1/8/5K2/8 b - - 3 4]
            \setchessboard{boardfontsize=12pt,labelfontsize=6pt,color=yellow!70,showmover=true,mover=b}
            \chessboard
            }
        \end{subfigure}
    \caption{
    (Left) After playing \variation{1. Rg6+} where both Rooks are left unprotected. (Right) Continuing with \variation{1...Kxf7 2. Qa1 hxg6 3. Qf6+ Kg8 4. Bh6!} avoiding the tempting \variation{4. Qxg6+}, after which victory slips away. White is completely winning at the end of the variation.}
    \label{fig:main_puzzle}
\end{figure}


\amatzia describes this as a ``short, yet challenging puzzle'' with an aesthetically pleasing `long' move, performing both attacking and defensive missions; while initiating a vicious threat, it continues to guard against the black Queen checks.




\matthew liked the retreats and the geometrical motifs (\variation{2. Qa1} \variation{3.Qf6+}), noting that they are very engine-like and tricky for humans to spot, requiring vision over the whole board. The touch of leaving the Rook on f7 \textit{en prise} while covering the check on d4 with the Queen on a1 is particularly fine.

\jonathan described this position as a very good combination involving the paradox (and limited depth) of the initial sacrifice, the geometry of the long queen move, and one or two nice variations before the final goal is established. The initial sacrifice needs to be accurately calculated, you need to see the long queen move (\variation{2.Qa1}), which is not at all typical and thus harder to see than a normal move.









\noindent {\large\textbf{Puzzles Highlighted by IM for Chess Compositions Amatzia Avni}}

\noindent \textbf{Puzzle~\ref{fig:puzzle_1}} (Left) shows the first highlighted puzzle. Here \variation{1... Bc3} requires a long calculation, since white has counter play, and the black king does not hesitate to move forward into the danger zone. Alternative lines of play including {\variation[invar]{1... e3 2. Qb5}} and {\variation[invar]{1... Qb1+ 2. Kxd2 Bc3+ 3. Kxc3 Qc1+ 4. Kb3 Qb1+ 5. Ka3}} appear fancy but are unproductive. The solution continues with \variation{2. Ne5+ Kf6 3. Rg6+ Kxf5 4. Qxc8+ Qxc8 5. Bg4+ Kxf4 6. Bxc8 Kxe5} leading to a win for black. This mainline of play is counter-intuitive and is not easy to assess beforehand.

\begin{figure}[h]
\centering
    \begin{subfigure}[t]{0.48\textwidth}
    \centering
    \adjustbox{max width=\textwidth}
        {
            \newchessgame[setfen=1qb5/5k2/P1Qp1bNp/2pP1P2/2P1pP1P/8/3rB1R1/4K3 b]
            \setchessboard{boardfontsize=12pt,labelfontsize=6pt,color=yellow!70,showmover=true,mover=b}
            \chessboard
        }
        \end{subfigure}
    ~
    \begin{subfigure}[t]{0.48\textwidth}
    \centering
    \adjustbox{max width=\textwidth}
        {
            \newchessgame[setfen= 2B5/8/P2p2Rp/2pPk3/2P1p2P/2b5/3r4/4K3 w - - 0 7]
            \setchessboard{boardfontsize=12pt,labelfontsize=6pt,color=yellow!70,showmover=true,mover=w}
            \chessboard
            }
        \end{subfigure}
    \caption{(Left) Puzzle position
    \textbf{\href{https://lichess.org/analysis/1qb5/5k2/P1Qp1bNp/2pP1P2/2P1pP1P/8/3rB1R1/4K3 b}{[Analyse on Lichess]}}. (Right) After playing \variation{1... Bc3 2. Ne5+ Kf6 3. Rg6+ Kxf5 4. Qxc8+ Qxc8 5. Bg4+ Kxf4 6. Bxc8 Kxe5}, which leads to a winning position for black.
    }
    \label{fig:puzzle_1}
\end{figure}


\noindent \textbf{Puzzle~\ref{fig:puzzle_2}} involves a slightly imbalanced position favoring black in terms of material advantage, though most of the material is not developed. The position begins with the white Rook on f7 hanging as it is under threat by the King on g8.

\begin{figure}[h]
\centering
    \begin{subfigure}[t]{0.48\textwidth}
    \centering
    \adjustbox{max width=\textwidth}
        {
            \newchessgame[setfen=rnbqrbk1/pp3Rp1/2p1p1N1/3p1P1Q/3PnB2/2P5/PP3P1P/6K1 w]
            \setchessboard{boardfontsize=12pt,labelfontsize=6pt,color=yellow!70,showmover=true,mover=w}
            \chessboard
        }
        \end{subfigure}
    ~
    \begin{subfigure}[t]{0.48\textwidth}
    \centering
    \adjustbox{max width=\textwidth}
        {
            \newchessgame[setfen= rnb1rb2/pp2q1p1/2p1pk2/3pNP1Q/3PnB2/2P5/PP3P1P/6K1 b - - 5 4]
            \setchessboard{boardfontsize=12pt,labelfontsize=6pt,color=yellow!70,showmover=true,mover=b}
            \chessboard
            }
        \end{subfigure}
    \caption{(Left) Puzzle position \textbf{\href{https://lichess.org/analysis/rnbqrbk1/pp3Rp1/2p1p1N1/3p1P1Q/3PnB2/2P5/PP3P1P/6K1 w}{[Analyse on Lichess]}}. (Right) Position after playing \variation{1. Re7! Qxe7 2. Qh8+ Kf7 3. Ne5+ Kf6 4. Qh5!!}. Continuing with \variation{4... Qc7 5. Qxe8}, white is winning.
    }
    \label{fig:puzzle_2}
\end{figure}


\mainline{1. Re7!} is the first move of the puzzle. This allows black to capture the Rook, though it removes an important flight square from the King. The solution continues with \mainline{1...Qxe7 2. Qh8+ Kf7 3. Ne5+ Kf6 4. Qh5!!} produces a quiet move to crown the combination, as shown in \Cref{fig:puzzle_2} (Right). Continuing with \mainline{4... Qc7 5. Qxe8}, where white is still down in material, but has a winning attack.

While black begins with enormous material advantage, the player is helpless in the above position. An alternative line with the obvious sequence of moves fails to deliver: {\variation[invar]{4. Qh4+ g5 5. fxg6+ Kg7 6. Qh7+ Kf6 7. Qh8+ Bg7??}} (After the correct {\variation[invar]{7...Qg7}} white must be satisfied with producing perpetual checks). Black succumbs to a devious trap with mate with two self-blocks: {\variation[invar]{8. Qh4+ Kf5 9. Qg4+ Kf6 10. Bg5+! Nxg5 11. Qf4\#}}. 


\noindent \textbf{Puzzle~\ref{fig:puzzle_3}} (Left) shows the next highlighted puzzle. The solution to the puzzle begins with \variation{1. Qe6 Bxe7 2.d8=N!}. The alternative line {\variation[invar]{1. d8=Q Rxd8 2. Qxd8 Qxe7}} loses. {\variation[invar]{2.Qxe7 Nxd7 3.Rxd7 Qxd7 4.Qxd7 Nb4}} is unclear; {\variation[invar]{2. d8=Q+}} is begged to be played after \variation{1.Qe6 Bxe7}, but the paradoxical response {\variation[invar]{2...Bf8!!}} will leave us shocked. The new Queen on d8 is lost, after which black gains the upper hand, for instance {\variation[invar]{3. Qf5 Rxd8 4.Rxd8 Qh1+ 5. Rd1 Qa8 6.Rf1 Bd6}}. The puzzle's solution continues with \variation{2...Qf3 3.Nf7+ Qxf7 4.Qxf7 Bf8 5.bxc4+}, leading to a win for white.

\begin{figure}[h]
\centering
    \begin{subfigure}[t]{0.48\textwidth}
    \centering
    \adjustbox{max width=\textwidth}
        {
            \newchessgame[setfen=r4b1k/pq1PN1pp/nn1Q4/2p3P1/2p4P/1Pp5/P7/2KR4 w]
            \setchessboard{boardfontsize=12pt,labelfontsize=6pt,color=yellow!70,showmover=true,mover=w}
            \chessboard
        }
        \end{subfigure}
    ~
    \begin{subfigure}[t]{0.48\textwidth}
    \centering
    \adjustbox{max width=\textwidth}
        {
            \newchessgame[setfen= r2N3k/pq2b1pp/nn2Q3/2p3P1/2p4P/1Pp5/P7/2KR4 b]
            \setchessboard{boardfontsize=12pt,labelfontsize=6pt,color=yellow!70,showmover=true,mover=b}
            \chessboard
            }
        \end{subfigure}
    \caption{(Left) Puzzle position \textbf{\href{https://lichess.org/analysis/r4b1k/pq1PN1pp/nn1Q4/2p3P1/2p4P/1Pp5/P7/2KR4 w}{[Analyse on Lichess]}}. (Right) After under-promoting the pawn with \variation{1. Qe6 Bxe7 2. d8=N!}, white eventually gains an upper hand.
    }
    \label{fig:puzzle_3}
\end{figure}

\newpage
\noindent {\large\textbf{Puzzles Highlighted by GM Jonathan Levitt}}

\noindent \textbf{Puzzle~\ref{fig:puzzle_4}} is a position that could easily be from a real game. White needs to find a fine sequence of exact moves to convert the point and the play flows across the board in a pleasing way. There are only minor aspects of paradox involved in the moves (for example capturing only a pawn on the first move when a piece is \textit{en prise}), but it would score reasonably well on geometry, flow and depth. All in all, this puzzle is very close to becoming endgame study material. A study composer might try to extend this a move or two at the beginning and introduce a strongly paradoxical sacrifice to set up the starting position…then it would have a little bit of everything. 

\begin{figure}[h]
\centering
    \begin{subfigure}[t]{0.48\textwidth}
    \centering
    \adjustbox{max width=\textwidth}
        {
            \newchessgame[setfen=8/4p3/p4p2/P1K2b2/BP1p2k1/2P1n3/3P2P1/8 w]
            \setchessboard{boardfontsize=12pt,labelfontsize=6pt,color=yellow!70,showmover=true,mover=w}
            \chessboard
        }
        \end{subfigure}
    ~
    \begin{subfigure}[t]{0.48\textwidth}
    \centering
    \adjustbox{max width=\textwidth}
        {
            \newchessgame[setfen=8/4p3/p4p2/P1K2b2/BP1P2k1/8/3P2n1/8 w - - 0 2]
            \setchessboard{boardfontsize=12pt,labelfontsize=6pt,color=yellow!70,showmover=true,mover=w}
            \chessboard
            }
        \end{subfigure}
    \caption{(Left) Puzzle position \textbf{\href{https://lichess.org/analysis/8/4p3/p4p2/P1K2b2/BP1p2k1/2P1n3/3P2P1/8 w}{[Analyse on Lichess]}}. (Right) After playing \variation{1. cxd4 Nxg2}, a counter-intuitive start to the variation.
    }
    \label{fig:puzzle_4}
\end{figure}

The mainline of play starts with \mainline{1. cxd4 Nxg2} which is not at all easy and involves rejecting the variations {\variation[invar]{1. Kxd4}} which is reasonable, but only draws with optimal play, and also {\variation[invar]{1. dxe3? dxc3 2. Kb6 Bd3}} where black is winning.


The solution to the puzzle continues with \mainline{2. Kb6 Bc8 3. Bc6 Nf4 4. Bb7 Bxb7 5. Kxb7 Nd5 6. b5 Nb4 7. bxa6 Nxa6 8. Kxa6 f5 9. Kb6 f4 10. a6 f3 11. a7 f2 12. a8=Q f1=Q 13. Qg8+} and white will pick up the e7 pawn with a winning position. 

\noindent \textbf{Puzzle~\ref{fig:puzzle_5}} is an elegant endgame position and is close to being endgame study standard. The position has a slight advantage to white in terms of material. The mainline of play is \variation{1... a2 2. Rd1 Bf5!}, which leads to black threatening with {\variation[invar]{3...Bb1}}, which would secure promotion. Now, {\variation[invar]{3. Re1 Kd8!}} prevents white from transferring the Rook behing the passed a-pawn. The solution continues with \variation{3. Rf1 Kd8!!}, which is an elegant little move that finishes white off. Note that black cannot immediately play {\variation[invar]{3... Bb1?}} due to {\variation[invar]{4. Rxf7+}}. After \variation{3... Kd8}, the line can continue with {\variation[invar]{4. Kc6 Bb1 5. Rxf7 Be4+!}} which leads to black promoting to a Queen. The solution continues with \variation{4. Ra1 Bb1}, where white is now completely paralyzed. Black has the simple plan of pushing the f and h pawns and wins this position without much effort.

\begin{figure}[h]
\centering
    \begin{subfigure}[t]{0.48\textwidth}
    \centering
    \adjustbox{max width=\textwidth}
        {
            \newchessgame[setfen=8/3k1p2/3Pb3/2K5/7p/p7/3R2P1/8 b]
            \setchessboard{boardfontsize=12pt,labelfontsize=6pt,color=yellow!70,showmover=true,mover=b}
            \chessboard
        }
        \end{subfigure}
    ~
    \begin{subfigure}[t]{0.48\textwidth}
    \centering
    \adjustbox{max width=\textwidth}
        {
            \newchessgame[setfen=8/3k1p2/3P4/2K2b2/7p/8/p5P1/3R4 w - - 2 3]
            \setchessboard{boardfontsize=12pt,labelfontsize=6pt,color=yellow!70,showmover=true,mover=w}
            \chessboard
            }
        \end{subfigure}
    \caption{(Left) Puzzle position \textbf{\href{https://lichess.org/analysis/8/3k1p2/3Pb3/2K5/7p/p7/3R2P1/8 b}{[Analyse on Lichess]}}. (Right) After playing \variation{1... a2 2. Rd1 Bf5!}, black is in position to threaten moving the Bishop to b1 on the next move.
    }
    \label{fig:puzzle_5}
\end{figure}

\noindent {\large\textbf{Puzzles Highlighted by GM Matthew Sadler}}

\noindent In \textbf{Puzzle~\ref{fig:puzzle_6}}, the key to the position is extremely nice. Within a couple of moves, a combination of themes was achieved which leaves a very elegant impression. This combination of an under-promotion with a smothered mate is not something I've seen before! 

\begin{figure}[h]
\centering
    \begin{subfigure}[t]{0.48\textwidth}
    \centering
    \adjustbox{max width=\textwidth}
        {
            \newchessgame[setfen=1q4rk/ppr1PQpp/1b3R2/3R4/1P6/4P3/P5PP/6K1 w]
            \setchessboard{boardfontsize=12pt,labelfontsize=6pt,color=yellow!70,showmover=true,mover=w}
            \chessboard
        }
        \end{subfigure}
    ~
    \begin{subfigure}[t]{0.48\textwidth}
    \centering
    \adjustbox{max width=\textwidth}
        {
            \newchessgame[setfen=3N2rk/ppr2Qpp/1b3R2/8/1P6/4P3/P5PP/6K1 b - - 0 2]
            \setchessboard{boardfontsize=12pt,labelfontsize=6pt,color=yellow!70,showmover=true,mover=b}
            \chessboard
            }
        \end{subfigure}
    \caption{(Left) Puzzle position \textbf{\href{https://lichess.org/analysis/1q4rk/ppr1PQpp/1b3R2/3R4/1P6/4P3/P5PP/6K1 w}{[Analyse on Lichess]}}. (Right) After under-promoting the pawn into a Knight with \variation{1. Rd8 Qxd8 2. exd8=N}.
    }
    \label{fig:puzzle_6}
\end{figure}

\variation{1. Rd8 Qxd8} seemingly a clever defense, exploiting the pin on the pawn on e7 by the rook on c7 as {\variation[invar]{1... Rxd8 2. Qf8+}}. However \variation{2. exd8=N} introduces a very unusual smothered mate motif,  which I don't think I've ever seen before combined with an under-promotion! The only shame is that black has a way to lead the game into a position where white still has to work hard for the win!
{\variation[invar]{2. exd8=Q}} is met by {\variation[invar]{2...Rxf7}}. The mainline of play continues with \variation{2... Rc1+ 3. Rf1 Bxe3+ 4. Kh1 Rxf1+ 5. Qxf1 Rxd8 6. g3 }. The engines assess this as a winning position for white.The position is a touch unnatural perhaps - the rook on c7 and queen on b8 make a poor impression - but not enough to cause any upset.


\noindent \textbf{Puzzle~\ref{fig:puzzle_7}} shows an extremely surprising puzzle where the threats and counter-threats against both Kings leave you in doubt as to the correct result until the very last move!

\begin{figure}[h]
\centering
    \begin{subfigure}[t]{0.48\textwidth}
    \centering
    \adjustbox{max width=\textwidth}
        {
            \newchessgame[setfen= 6rk/Q7/3q4/5p2/2PP1P2/P5Pr/7P/R4RK1 b]
            \setchessboard{boardfontsize=12pt,labelfontsize=6pt,color=yellow!70,showmover=true,mover=b}
            \chessboard
        }
        \end{subfigure}
    ~
    \begin{subfigure}[t]{0.48\textwidth}
    \centering
    \adjustbox{max width=\textwidth}
        {
            \newchessgame[setfen=6rk/Q7/8/5p2/2PP1P1q/P5P1/6K1/R4R2 w - - 3 4]
            \setchessboard{boardfontsize=12pt,labelfontsize=6pt,color=yellow!70,showmover=true,mover=w}
            \chessboard
            }
        \end{subfigure}
    \caption{(Left) Puzzle position \textbf{\href{https://lichess.org/analysis/6rk/Q7/3q4/5p2/2PP1P2/P5Pr/7P/R4RK1 b}{[Analyse on Lichess]}}. (Right) After playing the non-obvious line \variation{1... Rxh2 2. Kxh2 Qh6+ 3. Kg2 Qh4}.
    }
    \label{fig:puzzle_7}
\end{figure}


\variation{1... Rxh2 2. Kxh2 Qh6+ 3. Kg2 Qh4} is the correct non-obvious line. The nice point about this puzzle is how many "obviously winning" lines fail! \variation{4. Rf3} is the most obvious defense and seemingly ends the game! The line continues with \variation{4... Rxg3+ 5. Rxg3 Qh2+ 6. Kf3 Qf2+ 7. Kxf2} which leads to a very unexpected stalemate! The alternative line of play {\variation[invar]{4. Rh1}} involves pinning the Queen to the King, and will lose to {\variation[invar]{4...Rxg3+ 5. Kf2 Rh3+ 6. Ke2 Qg4+ 7. Kd2 Qg2+ 8. Kc1 Rxh1\#}}.

\noindent \textbf{Puzzle~\ref{fig:puzzle_8}} is a lovely combination of themes, adding a diversion to a typical smothered mate motif!  

\begin{figure}[h]
\centering
    \begin{subfigure}[t]{0.48\textwidth}
    \centering
    \adjustbox{max width=\textwidth}
        {
            \newchessgame[setfen= r4r1k/6pp/4Q3/5pN1/p2P1P2/1P5P/q5P1/2R4K w]
            \setchessboard{boardfontsize=12pt,labelfontsize=6pt,color=yellow!70,showmover=true,mover=w}
            \chessboard
        }
        \end{subfigure}
    ~
    \begin{subfigure}[t]{0.48\textwidth}
    \centering
    \adjustbox{max width=\textwidth}
        {
            \newchessgame[setfen=r1R2rk1/5Npp/4Q3/3P1p2/p4P2/1q5P/6P1/7K b - - 0 3]
            \setchessboard{boardfontsize=12pt,labelfontsize=6pt,color=yellow!70,showmover=true,mover=b}
            \chessboard
            }
        \end{subfigure}
    \caption{(Left) Puzzle position \textbf{\href{https://lichess.org/analysis/r4r1k/6pp/4Q3/5pN1/p2P1P2/1P5P/q5P1/2R4K w}{[Analyse on Lichess]}}. (Right) After playing \variation{1. Rc8 Qxb3 2. Nf7+ Kg8 3. d5}.
    }
    \label{fig:puzzle_8}
\end{figure}


\variation{1. Rc8 Qxb3} is the initial sequence of moves. The opponent has two other responses to the first move, each of which leads to a short mate sequence. Specifically, those alternative lines are {\variation[invar]{1... Rfxc8 2. Nf7+ Kg8 3. Nh6+ Kh8 4. Qg8+ Rxg8 5. Nf7\#}} and {\variation[invar]{1... Raxc8 2. Nf7+ Rxf7 3. Qxc8+}}, which ends with a back-rank mate. The puzzle solution continues with \variation{2. Nf7+ Kg8 3. d5}, which produces another neat addition to the puzzle as shown in Puzzle~\ref{fig:puzzle_8} (Right).


\variation{3...h6} is the most natural human defense in this position, but that is followed with \variation{4. Ne5+ Kh7 5. Qg6+ Kh8 6. Rc7}. The chess engine at this point estimates all lines as winning for the white player. 

\noindent {\large\textbf{Discussion}}

Our work demonstrates a significant advancement in the AI-driven generation of creative chess puzzles. We developed a puzzle generation approach with AI and produced a booklet of chess puzzles which, upon review by chess experts, received positive feedback. The positions in this booklet represent a pioneering step toward human-AI partnership in chess composition, and the experts who reviewed them highlighted several key qualities: they were beautiful, counter-intuitive, introduced novel variations on existing themes, and were aesthetically pleasing to solve. They particularly enjoyed puzzles with natural-looking positions and those that were original, paradoxical, and surprising. We also observed that the experts often selected different puzzles as their favorites, which suggests that creativity and beauty in chess are highly subjective.

Our work provides a new framework for discovering novel concepts, moving beyond known patterns and motifs. The methodology can be extended to support creative puzzle co-creation with human experts. Ultimately, we plan to generalize these results beyond chess, first to other board games and then to broader problem-solving domains.

\noindent {\large\textbf{Acknowledgements}}
We would like to thank Demis Hassabis for discussion and feedback on our work. We thank Aleksei Ostapenko, Arpit Hamirwasia, Daniel Körnlein, John Reid, Joon Lee, Kola Adeyemi, Matteo Tortora, Preet Sardhara, Sachin Ravichandran, Sagar Jha, and Temirlan Ulugbek uulu for their participation in evaluating our chess puzzles through various rating tasks, and particularly for their engagement in our Human study. We also thank them for their valuable discussions. The feedback from these tasks was instrumental in directing our work. We would also like to thank Lisa Schut for feedback on the booklet, David Abel for reviewing the paper, Simon Osindero, Arnaud Doucet and Clare Lyle for joining discussions and providing feedback, and Gabriela Fernandez-Cuervo for engaging with chess experts.

\newpage
\bibliographystyle{abbrvnat}
\bibliography{main}

\appendix
\begin{twocolumn}

\input{booklet}
\end{twocolumn}

\end{document}

%% file: booklet.tex
\section{Creative Chess Puzzles Booklet}
\label{sec:booklet}
The appendix presents the booklet of chess puzzles created using Artificial Intelligence (AI) techniques. This is a brief, non-technical summary of the methods used; for more detail, please refer to the paper.

\subsection{Sacrifice} 
\label{sec:themed_puzzles}

\noindent\textbf{Book example:}

\adjustbox{max width=\appendixchessboardwidth\columnwidth}{
    \chessboard[showmover=true, color=yellow!70,
    setfen= k1b5/pppN4/1R6/8/Q6K/8/8/8 w - - 0 1]
    }

\begin{center}
    \textbf{\href{https://lichess.org/analysis/k1b5/pppN4/1R6/8/Q6K/8/8/8 w - - 0 1}{[Analyse on Lichess]}}
\end{center}



\noindent\textbf{Selected puzzles:}

\adjustbox{max width=\appendixchessboardwidth\columnwidth}{
    \chessboard[
        showmover=true,
        color=yellow!70,
        setfen= r5k1/1b3NPp/p3r1n1/2pq4/3p4/8/PPP2QPP/4RRK1 w - - 0 1
    ]
    }
\begin{center}
\textbf{\href{https://lichess.org/analysis/r5k1/1b3NPp/p3r1n1/2pq4/3p4/8/PPP2QPP/4RRK1 w - - 0 1}{[Analyse on Lichess]}}

\textbf{Closest FENs - \href{https://lichess.org/analysis/r5k1/1bp3pp/p7/1p2q3/8/8/PPP2QPP/5RK1 w - - 0 23}{[1]}}, \textbf{\href{https://lichess.org/analysis/r5k1/6pp/p3p3/3pq3/8/8/PP3QPP/5RK1 w - - 0 33}{[2]}}, \textbf{\href{https://lichess.org/analysis/r5k1/1b3ppp/p2q1n2/2pN4/8/8/PPPQ1PPP/4R1K1 w - - 1 21}{[3]}}
\end{center}

1. Nd8! \chesscomments{The knight is sacrificed two ways, but cannot be captured due to Qf7\#. Black is forced to give up material.} Qxg2+ 2. Qxg2 Bxg2 3. Rf8+! \chesscomments{An intermezzo sacrifice that aims to liquidate.} Nxf8 4. gxf8=Q+ Kxf8 5. Nxe6+ Kf7 6. Kxg2 \chesscomments{White plays the endgame up a piece.}

\adjustbox{max width=\appendixchessboardwidth\columnwidth}
        {
            \chessboard[
                showmover=true,
                color=yellow!70,
                setfen= 1r1r2k1/Q2p1R1p/2p2R2/1p3pB1/1P4q1/8/5K2/8 w - - 0 1
            ]
        }
\begin{center}
    \textbf{\href{https://lichess.org/analysis/1r1r2k1/Q2p1R1p/2p2R2/1p3pB1/1P4q1/8/5K2/8 w - - 0 1}{[Analyse on Lichess]}}

\textbf{Closest FENs - \href{https://lichess.org/analysis/3r2k1/2p2p2/p2r3p/3PQ1PP/8/8/5P2/6K1 w - - 3 33}{[1]}}, \textbf{\href{https://lichess.org/analysis/3r2k1/8/p2R3p/1pn3p1/1P3p2/7P/5K2/8 w - - 0 50}{[2]}}, \textbf{\href{https://lichess.org/analysis/6k1/2R3p1/1p4Np/7P/4b1r1/8/5K2/8 w - - 2 35}{[3]}}
\end{center}

1. Rg6+! \chesscomments{White gives up both rooks to open up the a1-h8 diagonal. Capturing either rook eventually transposes to the same position.} hxg6 2. Qa1! Kxf7 3. Qf6+ Kg8 4. Bh6 \chesscomments{and White covers all the checks}.

\adjustbox{max width=\appendixchessboardwidth\columnwidth}
        {
            \chessboard[
                showmover=true,
                color=yellow!70,
                setfen= 5b1r/k2rqP2/BRR5/4P1p1/1n1N3p/8/3K1Q1P/1r6 w - - 0 1
            ]
        }
\begin{center}
    \textbf{\href{https://lichess.org/analysis/5b1r/k2rqP2/BRR5/4P1p1/1n1N3p/8/3K1Q1P/1r6 w - - 0 1}{[Analyse on Lichess]}}

\textbf{Closest FENs - \href{https://lichess.org/analysis/5n2/4kPK1/8/5P2/7p/8/7P/8 w - - 0 66}{[1]}}, \textbf{\href{https://lichess.org/analysis/8/2rR4/2n5/4kpp1/3N4/8/3K3P/8 w - - 0 45}{[2]}}, \textbf{\href{https://lichess.org/analysis/8/8/PR6/4k1p1/7p/7r/5K2/8 w - - 0 69}{[3]}}
\end{center}

1. Rb7+! \chesscomments{White gives up the rook to open the diagonal for the queen.} Rxb7 2. Nb5+ Kb8 3. Qa7+ \chesscomments{sacrificing the queen to finish the game.} Rxa7 4. Rc8\#.

\adjustbox{max width=\appendixchessboardwidth\columnwidth}
        {
            \chessboard[
                showmover=true,
                color=yellow!70,
                setfen= rq2rn1k/p1p1n1R1/bp5P/4qBp1/1P1p3N/PQP5/3K1P1P/7R w - - 0 1
            ]
        }
\begin{center}
    \textbf{\href{https://lichess.org/analysis/rq2rn1k/p1p1n1R1/bp5P/4qBp1/1P1p3N/PQP5/3K1P1P/7R w - - 0 1}{[Analyse on Lichess]}}

\textbf{Closest FENs - \href{https://lichess.org/analysis/rq2r1k1/1p1P1p1p/p1n5/4PBp1/1PQB3b/P7/5P1P/R3K2R b KQ - 0 26}{[1]}}, \textbf{\href{https://lichess.org/analysis/r3r1k1/p3q2p/1p4p1/5p1Q/1P1p4/P3P3/2PK1P2/6RR w - - 0 22}{[2]}}, \textbf{\href{https://lichess.org/analysis/r1b2r1k/p1p1R3/1p5P/4pp2/2P4N/P1q5/6P1/R5K1 w - - 0 27}{[3]}}
\end{center}

1. Qg8+ Nxg8 2. Rh7+ Nxh7 3. Ng6\# \chesscomments{A creative example of a smothered mate.}

\adjustbox{max width=\appendixchessboardwidth\columnwidth}
        {
            \chessboard[
                showmover=true,
                color=yellow!70,
                setfen= 7r/4R1R1/2r2p1p/p5pk/2p5/P1P3PP/1n5K/8 w - - 0 1
            ]
        }
\begin{center}
    \textbf{\href{https://lichess.org/analysis/7r/4R1R1/2r2p1p/p5pk/2p5/P1P3PP/1n5K/8 w - - 0 1}{[Analyse on Lichess]}}

\textbf{Closest FENs - \href{https://lichess.org/analysis/8/r7/4R2p/pp4pk/2p5/P1P2KPP/8/8 w - - 0 42}{[1]}}, \textbf{\href{https://lichess.org/analysis/8/8/7p/p4Kpk/8/P5PP/8/8 w - - 0 44}{[2]}}, \textbf{\href{https://lichess.org/analysis/8/8/3k2p1/pp5p/2pK1P2/P1P3P1/7P/8 w - - 2 38}{[3]}}
\end{center}

1. Re4 \chesscomments{Setting up a mating net.} f5 2. Rh4+! gxh4 3. g4+ fxg4 4. hxg4\#. 

\vfill\null

\subsection{Underpromotion \citep{spectacularchess}}

\noindent\textbf{Book example:}

\adjustbox{max width=\appendixchessboardwidth\columnwidth}
        {
            \chessboard[
                showmover=true,
                color=yellow!70,
                setfen= 8/1KP5/8/2p5/1pP5/p7/k7/1R3R2 w - - 0 1
            ]
            }
\begin{center}
    \textbf{\href{https://lichess.org/analysis/8/1KP5/8/2p5/1pP5/p7/k7/1R3R2 w - - 0 1}{[Analyse on Lichess]}}
\end{center}

\noindent\textbf{Selected puzzles:}

\adjustbox{max width=\appendixchessboardwidth\columnwidth}
        {
            \chessboard[
                showmover=true,
                color=yellow!70,
                setfen= QR3nk1/p4pp1/7p/6q1/8/2PB3P/r4p1K/5R2 b - - 0 1
            ]
            }
\begin{center}
    \textbf{\href{https://lichess.org/analysis/QR3nk1/p4pp1/7p/6q1/8/2PB3P/r4p1K/5R2 b - - 0 1}{[Analyse on Lichess]}}

\textbf{Closest FENs - \href{https://lichess.org/analysis/6k1/5pp1/8/3p2q1/8/2PB3P/r4P2/4RK2 w - - 3 39}{[1]}}, \textbf{\href{https://lichess.org/analysis/1R3nk1/p5p1/7p/6q1/4Q3/7P/5PP1/6K1 b - - 4 32}{[2]}}, \textbf{\href{https://lichess.org/analysis/6k1/p5p1/3N3p/8/8/2PR3P/r6r/2R2K2 b - - 6 29}{[3]}}
\end{center}

1... Qg1+! \chesscomments{Black gives up the queen to set up the underpromotion and mate.} 2. Rxg1 f1=N+! 3. Kh1 Rh2\#.

\adjustbox{max width=\appendixchessboardwidth\columnwidth}
        {
            \chessboard[
                showmover=true,
                color=yellow!70,
                setfen= 8/P7/8/2bP4/2p2k2/4p3/4Kp2/5N2 b - - 0 1
            ]
        }
\begin{center}
    \textbf{\href{https://lichess.org/analysis/8/P7/8/2bP4/2p2k2/4p3/4Kp2/5N2 b - - 0 1}{[Analyse on Lichess]}}

\textbf{Closest FENs - \href{https://lichess.org/analysis/8/8/B7/8/6k1/5p2/5Kp1/8 b - - 13 73}{[1]}}, \textbf{\href{https://lichess.org/analysis/8/8/8/8/2pk4/3p4/3Kp3/R7 b - - 1 56}{[2]}}, \textbf{\href{https://lichess.org/analysis/8/P7/8/3N4/2p3k1/2P2p2/6p1/6K1 b - - 0 54}{[3]}}
\end{center}

1... Bxa7 2. d6 c3 3. d7 \chesscomments{The position looks like it is heading for a draw as it seems like the black bishop has to stop White’s pawn. Black however has another idea in mind.} c2! 4. d8=Q c1=N! \chesscomments{The only winning line! Black underpromotes with tempo.} 5. Kd1 e2+ 6. Kc2 exf1=Q \chesscomments{Black can eventually escape the checks finding shelter on g1 covered by the new queen.}

\adjustbox{max width=\appendixchessboardwidth\columnwidth}
        {
            \chessboard[
                showmover=true,
                color=yellow!70,
                setfen= r4b1k/pq1PN1pp/nn1Q4/2p3P1/2p4P/1Pp5/P7/2KR4 w - - 0 1
            ]
        }
\begin{center}
    \textbf{\href{https://lichess.org/analysis/r4b1k/pq1PN1pp/nn1Q4/2p3P1/2p4P/1Pp5/P7/2KR4 w - - 0 1}{[Analyse on Lichess]}}

\textbf{Closest FENs - \href{https://lichess.org/analysis/4r1k1/p2P1ppp/q7/1p3P2/6P1/P1p4P/1PP5/2KR4 w - - 0 33}{[1]}}, \textbf{\href{https://lichess.org/analysis/7k/pp2r1pp/8/6P1/7P/1P6/P7/1K1R4 w - - 0 38}{[2]}}, \textbf{\href{https://lichess.org/analysis/r5k1/r1p2ppp/8/3RpP2/1p4P1/1P5P/2P5/2KR4 w - - 0 29}{[3]}}
\end{center}

1. Qe6 \chesscomments{Threatening mate.} Bxe7 2. d8=N! \chesscomments{Underpromoting to threaten a smothered mate. White's best option is to give up material.} Qf3 3. Nf7+ Qxf7 4. Qxf7

\adjustbox{max width=\appendixchessboardwidth\columnwidth}
        {
            \chessboard[
                showmover=true,
                color=yellow!70,
                setfen= 1q4rk/ppr1PQpp/1b3R2/3R4/1P6/4P3/P5PP/6K1 w - - 0 1
            ]
        }
\begin{center}
    \textbf{\href{https://lichess.org/analysis/1q4rk/ppr1PQpp/1b3R2/3R4/1P6/4P3/P5PP/6K1 w - - 0 1}{[Analyse on Lichess]}}

\textbf{Closest FENs - \href{https://lichess.org/analysis/7k/p1q3pp/8/8/1P6/4P3/P5PP/5RK1 w - - 0 40}{[1]}}, \textbf{\href{https://lichess.org/analysis/1r5k/p1p2Qpp/4p3/8/4q3/4P3/P5PP/5RK1 w - - 2 27}{[2]}}, \textbf{\href{https://lichess.org/analysis/5r1k/p3Q1pp/1q6/3p4/8/4P3/6PP/6K1 w - - 0 29}{[3]}}
\end{center}

1. Rd8! \chesscomments{White exploits Black's back-rank issues.} Qxd8! 2. exd8=N! \chesscomments{Underpromoting to a knight is the only move that wins!} Rc1+ 3. Kf2 Bxd8 4. Re6 \chesscomments{White is up a material with a winning position.}

\adjustbox{max width=\appendixchessboardwidth\columnwidth}
        {
            \chessboard[
                showmover=true,
                color=yellow!70,
                setfen= 1Q6/p1P1k3/PPqn4/2p1p1p1/4rp1P/2P3r1/5KP1/1R5R w - - 0 1
            ]
        }
\begin{center}
    \textbf{\href{https://lichess.org/analysis/1Q6/p1P1k3/PPqn4/2p1p1p1/4rp1P/2P3r1/5KP1/1R5R w - - 0 1}{[Analyse on Lichess]}}

\textbf{Closest FENs - \href{https://lichess.org/analysis/3Q4/p4k2/1P2p3/5p1p/7P/2P5/5KP1/1q6 b - - 2 33}{[1]}}, \textbf{\href{https://lichess.org/analysis/4n3/4k3/P2p4/2pKp1p1/5p2/1PP5/5PP1/8 w - - 1 42}{[2]}}, \textbf{\href{https://lichess.org/analysis/8/pp3P2/2k5/2p1p2p/5r2/P7/6KP/4R3 w - - 1 41}{[3]}}
\end{center}

1. c8=N! \chesscomments{White underpromotes to a knight.} Nxc8 \chesscomments{if 1... Qxc8 2. Qxc8 Nxc8 3. b7 or if 1... Ke6 2. Qxd6 Qxd6 3. Nxd6} 2. Qb7+! \chesscomments{Forcing matters.} Kd6 3. Qxc6 Kxc6 4. b7.

\adjustbox{max width=\appendixchessboardwidth\columnwidth}
        {
            \chessboard[
                color=yellow!70,
                setfen= rr4k1/1R3p2/4pPp1/3pPP1p/3qn1PK/8/6B1/2Q2R2 w - - 0 1
            ]
        }
\begin{center}
    \textbf{\href{https://lichess.org/analysis/rr4k1/1R3p2/4pPp1/3pPP1p/3qn1PK/8/6B1/2Q2R2 w - - 0 1}{[Analyse on Lichess]}}

\textbf{Closest FENs - \href{https://lichess.org/analysis/5k2/1R3p2/4pKp1/4P2p/5P2/6r1/8/8 w - - 0 67}{[1]}}, \textbf{\href{https://lichess.org/analysis/8/1R3p2/4kPp1/4P2p/6Pr/5K2/8/8 w - - 2 45}{[2]}}, \textbf{\href{https://lichess.org/analysis/r7/P3p3/3pP3/3Pk1p1/6K1/R7/8/8 w - - 3 50}{[3]}}
\end{center}

1. Rxb8+ Rxb8 2. fxg6 Nd2 \chesscomments{If 2... fxg6 3. Qc7 wins.} 3. gxf7+ Kh7! 4. f8=N+! Kh6! 5. Be4! \chesscomments{An underpromotion followed by a stunning bishop sacrifice. Black cannot capture the bishop.} Qe3 6. Nxe6 \chesscomments{White ends up with a large material advantage and should convert quickly with careful play.}


\subsection{Attacking Withdrawal~\citep{creativechess}}

\noindent\textbf{Book example:}

\adjustbox{max width=\appendixchessboardwidth\columnwidth}
        {
            \chessboard[
                showmover=true,
                color=yellow!70,
                setfen= 2r1rbk1/p1p2pp1/1p2p2p/n3Nq2/b1PPQB1P/2PB2P1/P4P2/1R2R1K1 w - - 0 1
            ]
            }
\begin{center}
    \textbf{\href{https://lichess.org/analysis/2r1rbk1/p1p2pp1/1p2p2p/n3Nq2/b1PPQB1P/2PB2P1/P4P2/1R2R1K1 w - - 0 1}{[Analyse on Lichess]}}
\end{center}



\newpage

\noindent\textbf{Selected puzzles:}

\adjustbox{max width=\appendixchessboardwidth\columnwidth}
        {
            \chessboard[
                showmover=true,
                color=yellow!70,
                setfen= 6rk/1b2q1pp/p2pBpRQ/3PpP2/2p1P3/2P4P/P7/6K1 w - - 0 1
            ]
            }
\begin{center}
    \textbf{\href{https://lichess.org/analysis/6rk/1b2q1pp/p2pBpRQ/3PpP2/2p1P3/2P4P/P7/6K1 w - - 0 1}{[Analyse on Lichess]}}

\textbf{Closest FENs - \href{https://lichess.org/analysis/6k1/p2q1ppp/2p1p3/2Pp4/1Q1P4/P3PPP1/6KP/8 w - - 0 27}{[1]}}, \textbf{\href{https://lichess.org/analysis/6k1/3r1p1p/2pNpQp1/2P1P3/1p6/1q4P1/5P2/6K1 w - - 2 41}{[2]}}, \textbf{\href{https://lichess.org/analysis/6k1/p4ppp/2p5/2Pp4/1P2q3/P6P/8/4Q1K1 w - - 3 34}{[3]}}
\end{center}

1. Rg4 \chesscomments{White retreats the rook and sets up the threat of 2. Qxh7+ Kxh7 3. Rh4\#.} g5 2. Bxg8 Kxg8 3. h4 \chesscomments{and White is quickly breaking through.}

\adjustbox{max width=\appendixchessboardwidth\columnwidth}
        {
            \chessboard[
                showmover=true,
                color=yellow!70,
                setfen= kr3bn1/7r/P2p4/Q2Pp1pp/1N2Pp2/6P1/4q2P/RR5K w - - 0 1
            ]
        }
\begin{center}
    \textbf{\href{https://lichess.org/analysis/kr3bn1/7r/P2p4/Q2Pp1pp/1N2Pp2/6P1/4q2P/RR5K w - - 0 1}{[Analyse on Lichess]}}

{\textbf{Closest FENs - \href{https://lichess.org/analysis/5bk1/7p/P2p1r2/3Pp1p1/1qpNPp2/5P2/6PP/1R5K w - - 0 41}{[1]}}, \textbf{\href{https://lichess.org/analysis/8/8/2p5/3k1pp1/N2Pp3/4K1P1/1r5P/8 w - - 0 51}{[2]}}, \textbf{\href{https://lichess.org/analysis/1k6/8/3K4/3Pp1pp/4Pp2/5P2/6nP/8 w - - 0 43}{[3]}}}
\end{center}

1. Nc2! \chesscomments{The white knight retreats and sacrifices itself to gain valuable time.} Qxe4+ 2. Kg1 Qxc2 3. Rxb8+ Kxb8 4. a7+ Rxa7  5. Qxa7+ \chesscomments{and White has an unstoppable attack.}

\adjustbox{max width=\appendixchessboardwidth\columnwidth}
        {
            \chessboard[
                showmover=true,
                color=yellow!70,
                setfen= r5rk/2Q2Rb1/1p4Bb/p7/P1P5/2q5/6PP/5R1K w - - 0 1
            ]
        }
\begin{center}
    \textbf{\href{https://lichess.org/analysis/r5rk/2Q2Rb1/1p4Bb/p7/P1P5/2q5/6PP/5R1K w - - 0 1}{[Analyse on Lichess]}}

\textbf{Closest FENs - \href{https://lichess.org/analysis/r4rk1/3Q1Rp1/1p5p/p1p5/2P5/2q5/6PP/5R1K w - - 0 23}{[1]}}, \textbf{\href{https://lichess.org/analysis/r6k/5R1p/p5pB/4b3/P1r5/8/6PP/5R1K w - - 0 26}{[2]}}, \textbf{\href{https://lichess.org/analysis/r6k/2p2Qpp/1p6/p7/P1P5/2q1P3/6PP/5RK1 w - - 1 26}{[3]}}
\end{center}

1. Bb1! \chesscomments{White withdraws the bishop from the attack with the goal of setting up a bishop-queen battery. It turns out that this plan is largely unstoppable.} Qb4 \chesscomments{Engine also gives Bg5, but after 2. R7f3 Black has to depart with the queen and is defending with less material.} 2. Qc6 \chesscomments{Threatening both 3. Qg6 and 3. Qxh6!. Black cannot defend.}  Rgd8 3. Qg6 Qxb1 \chesscomments{Black has to give up the queen.}



\adjustbox{max width=\appendixchessboardwidth\columnwidth}
        {
            \chessboard[
                showmover=true,
                color=yellow!70,
                setfen= 5rk1/1p2Rppp/p2Q1P2/8/8/7P/q4rP1/3R2K1 w - - 0 1
            ]
        }
\begin{center}
    \textbf{\href{https://lichess.org/analysis/5rk1/1p2Rppp/p2Q1P2/8/8/7P/q4rP1/3R2K1 w - - 0 1}{[Analyse on Lichess]}}

\textbf{Closest FENs - \href{https://lichess.org/analysis/5rk1/p2RQppp/8/8/3P4/7P/q4PP1/6K1 w - - 1 26}{[1]}}, \textbf{\href{https://lichess.org/analysis/5rk1/2R2ppp/p2Q4/4P3/8/8/q4rPP/3R2K1 w - - 0 25}{[2]}}, \textbf{\href{https://lichess.org/analysis/5rk1/1R2Qppp/p7/8/8/8/q4rPP/4R1K1 w - - 0 20}{[3]}}
\end{center}

1. Re3! \chesscomments{The onyl winning withdrawing move.} Rxg2+ 2. Kh1 gxf6 \chesscomments{If 2... h6 3. Qxf8+!} 3. Rg3+ Rxg3 4. Qxg3+ Kh8 5. Qd6! \chesscomments{Black cannot defend the hanging rook and pawn on f6.}



\subsection{Knight on the Rim is Dim~\citep{tigerchess}}

\noindent\textbf{Book example:}

\adjustbox{max width=\appendixchessboardwidth\columnwidth}
        {
            \chessboard[
                showmover=true,
                color=yellow!70,
                setfen= r1bqk2r/ppn1b1pp/2n2p2/2p1p3/8/1PN2NP1/PB1PPPBP/2RQ1RK1 w kq - 0 11
            ]
            }
\begin{center}
    \textbf{\href{https://lichess.org/analysis/r1bqk2r/ppn1b1pp/2n2p2/2p1p3/8/1PN2NP1/PB1PPPBP/2RQ1RK1 w kq - 0 11}{[Analyse on Lichess]}}
\end{center}


\noindent\textbf{Selected puzzles:}

\adjustbox{max width=\appendixchessboardwidth\columnwidth}
        {
            \chessboard[
                showmover=true,
                color=yellow!70,
                setfen= 4r1k1/4r1bp/1p5q/1Npp2N1/p1nP2Q1/P5B1/1PK3PP/3RR3 b - - 0 1
            ]
            }
\begin{center}
    \textbf{\href{https://lichess.org/analysis/4r1k1/4r1bp/1p5q/1Npp2N1/p1nP2Q1/P5B1/1PK3PP/3RR3 b - - 0 1}{[Analyse on Lichess]}}

\textbf{Closest FENs - \href{https://lichess.org/analysis/4r1k1/6p1/7p/p2q1p2/P2Q4/1P5P/5PP1/3R2K1 b - - 0 28}{[1]}}, \textbf{\href{https://lichess.org/analysis/3r2k1/6bp/1p4p1/3bp3/p3N3/P5P1/1P4KP/3RR3 b - - 5 28}{[2]}}, \textbf{\href{https://lichess.org/analysis/4k3/5bpp/1p6/1Npn4/6P1/P7/P1K4P/3R4 b - - 1 29}{[3]}}
\end{center}

1... Qg6+ 2. Kc1 Na5! \chesscomments{Black places the knight on the rim, threatening checkmate.} 3. b3 axb3 \chesscomments{White surprisingly has no good way to defend the threat of Qc2}.

\adjustbox{max width=\appendixchessboardwidth\columnwidth}
        {
            \chessboard[
                showmover=true,
                color=yellow!70,
                setfen= r4r1k/pp1p2R1/n2p3P/q1pPpN2/4P1P1/1Pb1PQP1/5n2/6K1 w - - 0 1
x            ]
        }
\begin{center}
    \textbf{\href{https://lichess.org/analysis/r4r1k/pp1p2R1/n2p3P/q1pPpN2/4P1P1/1Pb1PQP1/5n2/6K1 w - - 0 1}{[Analyse on Lichess]}}

\textbf{Closest FENs - \href{https://lichess.org/analysis/r4r1k/pp3R2/2p4P/3pN3/3P1P2/2P5/P7/K4n2 w - - 2 29}{[1]}}, \textbf{\href{https://lichess.org/analysis/r3r1k1/p4R1p/2p3pB/3pN3/3Pn2q/P2QP3/2P2P2/6K1 w - - 1 22}{[2]}}, \textbf{\href{https://lichess.org/analysis/r6k/p4R1n/2p1P2q/1pPpP1r1/1P1P4/P2NQ3/5RP1/6K1 b - - 2 39}{[3]}}
\end{center}

\adjustbox{max width=\appendixchessboardwidth\columnwidth}
        {
            \chessboard[
                showmover=true,
                color=yellow!70,
                setfen= 1rbqr1k1/r2p1Rb1/npp3N1/2P1p1pp/PP2Q3/8/4N1KP/5R2 w - - 0 1
            ]
        }
\begin{center}
    \textbf{\href{https://lichess.org/analysis/1rbqr1k1/r2p1Rb1/npp3N1/2P1p1pp/PP2Q3/8/4N1KP/5R2 w - - 0 1}{[Analyse on Lichess]}}

\textbf{Closest FENs - \href{https://lichess.org/analysis/1rb2r1k/4R2p/p7/1p1p1pPP/3P1P2/8/P7/5RK1 w - - 0 31}{[1]}}, \textbf{\href{https://lichess.org/analysis/2bqr2k/3p1Q1p/p1P3p1/1P1pP3/P7/8/6PP/5R1K w - - 1 34}{[2]}}, \textbf{\href{https://lichess.org/analysis/1rbqr1k1/2p3bp/6P1/p1pPp3/2n1N3/6Q1/PP4BP/R4R1K w - - 1 24}{[3]}}
\end{center}


\newpage
\adjustbox{max width=\appendixchessboardwidth\columnwidth}
        {
            \chessboard[
                showmover=true,
                color=yellow!70,
                setfen= 3B4/1R1P1pk1/6pb/r2Pp2b/3qPn1P/P2n4/3N2BK/2RQ4 b - - 0 1
            ]
        }
\begin{center}
    \textbf{\href{https://lichess.org/analysis/3B4/1R1P1pk1/6pb/r2Pp2b/3qPn1P/P2n4/3N2BK/2RQ4 b - - 0 1}{[Analyse on Lichess]}}

\textbf{Closest FENs - \href{https://lichess.org/analysis/2Q5/6pk/7p/5p2/4q3/1PP2n2/8/R2K4 b - - 7 68}{[1]}}, \textbf{\href{https://lichess.org/analysis/8/5pk1/6p1/4p2p/2Qq3P/P7/6PK/8 w - - 2 38}{[2]}}, \textbf{\href{https://lichess.org/analysis/8/2R2pk1/6p1/3Bn2p/4PP1P/3r4/6PK/8 b - - 2 42}{[3]}}
\end{center}



\subsection{Sacrifice Pieces to Stalemate~\citep{creativechess}}

\noindent\textbf{Book example:}

\adjustbox{max width=\appendixchessboardwidth\columnwidth}
        {
            \chessboard[
                showmover=true,
                color=yellow!70,
                setfen= 4Q3/6k1/p2pn2p/6p1/2q2pP1/5B1K/7P/8 w - - 0 1
            ]
            }
\begin{center}
    \textbf{\href{https://lichess.org/analysis/4Q3/6k1/p2pn2p/6p1/2q2pP1/5B1K/7P/8 w - - 0 1}{[Analyse on Lichess]}}
\end{center}


\clearpage
\noindent\textbf{Selected puzzles:}

\adjustbox{max width=\appendixchessboardwidth\columnwidth}
        {
            \chessboard[
                showmover=true,
                color=yellow!70,
                setfen= 8/2Q5/5qpk/6p1/2n3B1/P5P1/1P2RPK1/3r4 b - - 0 1
            ]
            }
\begin{center}
    \textbf{\href{https://lichess.org/analysis/8/2Q5/5qpk/6p1/2n3B1/P5P1/1P2RPK1/3r4 b - - 0 1}{[Analyse on Lichess]}}

\textbf{Closest FENs - \href{https://lichess.org/analysis/8/8/5qpk/7p/8/5Q1P/6PK/8 b - - 5 49}{[1]}}, \textbf{\href{https://lichess.org/analysis/8/8/6pk/7p/5p1P/1r4P1/4RPK1/8 b - - 1 59}{[2]}}, \textbf{\href{https://lichess.org/analysis/8/5p2/4pk2/7p/5p1P/1r4P1/1P2RPK1/8 b - - 1 37}{[3]}}
\end{center}

1... Ne3+! 2. Rxe3 \chesscomments{White is forced to accept the sacrifice due to the bishop hanging on g4.} Rg1+! \chesscomments{A second sacrifice!} 3. Kxg1 \chesscomments{(3. Kh3 Qxf2 and White is forced to give up the rook with Qb7 to stop mate.)} Qxf2+ 4. Kh1 Qg1+ 5. Kxg1 \chesscomments{Stalemate}.

\adjustbox{max width=\appendixchessboardwidth\columnwidth}
        {
            \chessboard[
                showmover=true,
                color=yellow!70,
                setfen= 6rk/2R1R1b1/p2Qp1Bb/P2pP2p/2pP3P/2q3PK/1r6/8 w
            ]
        }
\begin{center}
    \textbf{\href{https://lichess.org/analysis/6rk/2R1R1b1/p2Qp1Bb/P2pP2p/2pP3P/2q3PK/1r6/8 w - - 0 1}{[Analyse on Lichess]}}

\textbf{Closest FENs - \href{https://lichess.org/analysis/6rk/8/2Q2b2/1p6/p4P1P/1q4PK/8/4R3 w - - 2 38}{[1]}}, \textbf{\href{https://lichess.org/analysis/6rk/8/3Qp1q1/2pP2Bp/7P/2P3PK/Pr6/5R2 b - - 0 32}{[2]}}, \textbf{\href{https://lichess.org/analysis/6rk/6b1/p4R1p/1pp5/5Q1P/2q3PK/8/8 w - - 2 41}{[3]}}
\end{center}

1. Rxg7! \chesscomments{White starts a chain of sacrifices that surprisingly force a draw.}  Bxg7 \chesscomments{(1... Rxg7?? 2. Qf8+! Rg8 3. Qxh6\#)} 2. Rxg7!  Kxg7 3. Qe7+ Kxg6 4. Qh7+ Kxh7 \chesscomments{Stalemate.}

\adjustbox{max width=\appendixchessboardwidth\columnwidth}
        {
            \chessboard[
                showmover=true,
                color=yellow!70,
                setfen= 6rk/Q7/3q4/5p2/2PP1P2/P5Pr/7P/R4RK1 b
            ]
        }
\begin{center}
    \textbf{\href{https://lichess.org/analysis/6rk/Q7/3q4/5p2/2PP1P2/P5Pr/7P/R4RK1 b - - 0 1}{[Analyse on Lichess]}}

\textbf{Closest FENs - \href{https://lichess.org/analysis/6k1/8/2q2Q2/4p1P1/8/6Pp/7P/5RK1 b - - 0 55}{[1]}}, \textbf{\href{https://lichess.org/analysis/2r2rk1/Q5p1/4p2p/4P2q/3P1P2/P5P1/7P/R4RK1 b - - 0 33}{[2]}}, \textbf{\href{https://lichess.org/analysis/6k1/7p/3q4/8/1P4PK/5P2/4r2P/5RQ1 b - - 4 50}{[3]}}
\end{center}

1... Rxh2! 2. Kxh2 Qh6+ 3. Kg2 Qh4! \chesscomments{Black sets up strong mate threats that White has to address.} 4. Rf3 \chesscomments{(4. Rh1?? Rxg3 5. Kf2 Rh3+ and Black wins.)}  Rxg3+! 5. Rxg3 \chesscomments{Black has managed to set up the stalemate with the help of White’s rook.} Qh2+! 6. Kf3 Qe2+ 7. Kxe2 \chesscomments{Stalemate.}

\adjustbox{max width=\appendixchessboardwidth\columnwidth}
        {
            \chessboard[
                showmover=true,
                color=yellow!70,
                setfen= 3R4/6rk/5Q2/4P2P/5p2/pP3P2/2P5/1KN1q3 b - - 0 1
            ]
        }
\begin{center}
    \textbf{\href{https://lichess.org/analysis/3R4/6rk/5Q2/4P2P/5p2/pP3P2/2P5/1KN1q3 b - - 0 1}{[Analyse on Lichess]}}

\textbf{Closest FENs - \href{https://lichess.org/analysis/3Q4/p3P1rk/8/7P/5p2/1Pp5/P1P5/1K6 b - - 0 46}{[1]}}, \textbf{\href{https://lichess.org/analysis/3Q4/5rk1/3P4/4r3/4p3/P7/1P6/1K6 b - - 0 45}{[2]}}, \textbf{\href{https://lichess.org/analysis/8/8/8/4k2P/6P1/pp3P2/8/1K6 b - - 0 53}{[3]}}
\end{center}

\vfill\null
\adjustbox{max width=\appendixchessboardwidth\columnwidth}
        {
            \chessboard[
                showmover=true,
                color=yellow!70,
                setfen= 5rk1/R2Q1p1p/6pP/p2R1pP1/5P1K/r1q5/8/8 w - - 0 1
            ]
        }
\begin{center}
    \textbf{\href{https://lichess.org/analysis/5rk1/R2Q1p1p/6pP/p2R1pP1/5P1K/r1q5/8/8 w - - 0 1}{[Analyse on Lichess]}}

\textbf{Closest FENs - \href{https://lichess.org/analysis/6k1/5p1p/6p1/p2R1bPP/4r3/K7/8/8 w - - 1 32}{[1]}}, \textbf{\href{https://lichess.org/analysis/6k1/pQ6/2P4p/6pP/6P1/K1q5/8/8 w - - 24 94}{[2]}}, \textbf{\href{https://lichess.org/analysis/7k/1R5p/6pP/p4pP1/3p1P2/4r3/5K2/8 w - - 1 46}{[3]}}
\end{center}


\subsection{Novotny~\citep{spectacularchess}}

\noindent\textbf{Book example:}

\adjustbox{max width=\appendixchessboardwidth\columnwidth}
        {
            \chessboard[
                showmover=true,
                color=yellow!70,
                setfen= 2K1k3/1R4BN/3R4/5Nn1/8/2b5/4r3/3r4 w - - 0 1
            ]
            }
\begin{center}
    \textbf{\href{https://lichess.org/analysis/2K1k3/1R4BN/3R4/5Nn1/8/2b5/4r3/3r4 w - - 0 1}{[Analyse on Lichess]}}
\end{center}



\newpage
\noindent\textbf{Selected puzzles:}

\adjustbox{max width=\appendixchessboardwidth\columnwidth}
        {
            \chessboard[
                showmover=true,
                color=yellow!70,
                setfen= r4r1k/6pp/4Q3/5pN1/p2P1P2/1P5P/q5P1/2R4K w
            ]
            }
\begin{center}
    \textbf{\href{https://lichess.org/analysis/r4r1k/6pp/4Q3/5pN1/p2P1P2/1P5P/q5P1/2R4K w - - 0 1}{[Analyse on Lichess]}}

\textbf{Closest FENs - \href{https://lichess.org/analysis/b4r1k/6pp/4Q3/5n2/3b4/1P5P/P2R2P1/7K b - - 4 39}{[1]}}, \textbf{\href{https://lichess.org/analysis/1r5k/6pp/4Q3/1q1p1p2/5P2/1P6/6PP/2R4K w - - 0 32}{[2]}}, \textbf{\href{https://lichess.org/analysis/r1R1r1k1/5pp1/7p/4pN2/1P2n3/P3P2P/6P1/2R4K b - - 2 35}{[3]}}
\end{center}

1. Rc8! \chesscomments{White sacrifices the rook two ways!} Qxb3 \chesscomments{(1... Raxc8 Nf7+ and similarly 1... Rfxc8 Nf7+)} 2. Nf7+ d5! \chesscomments{Blocking the queen trade.} 3. Qd1+ Kh2 4. Qh5 Nd8+! 5. Kh8 Rxa8 \chesscomments{White is up a rook and quickly ending the game.}

\adjustbox{max width=\appendixchessboardwidth\columnwidth}
        {
            \chessboard[
                showmover=true,
                color=yellow!70,
                setfen= 4r1rk/p2Q2pp/2p1p3/3pP3/3Pq3/BP4R1/P3b1PP/5RK1 w
            ]
        }
\begin{center}
    \textbf{\href{https://lichess.org/analysis/4r1rk/p2Q2pp/2p1p3/3pP3/3Pq3/BP4R1/P3b1PP/5RK1 w - - 0 1}{[Analyse on Lichess]}}

\textbf{Closest FENs - \href{https://lichess.org/analysis/4r1k1/p2Q2pp/1p6/2p5/2Ppq3/P6P/1n4P1/5RK1 w - - 0 29}{[1]}}, \textbf{\href{https://lichess.org/analysis/4r2k/ppp2Qpp/3p4/3P4/2P1q3/1P6/P5PP/5RK1 w - - 6 24}{[2]}}, \textbf{\href{https://lichess.org/analysis/5rk1/p5pp/2pb4/3p4/3P4/1P6/P3N1PP/5RK1 b - - 1 23}{[3]}}
\end{center}

1. Bf8! \chesscomments{White sacrifices the bishop two ways!} g6 \chesscomments{(1...Rexf8 2. Qxg7+! Rxg7 3. Rxf8+ Rg8 Rfxg8\#)} 2. Qxe8 Bxf1 3. Qf7! Qf5 4. Rf3! Qxf7 5. Rxf7 \chesscomments{Black ends the combination with a dominating position.}


\adjustbox{max width=\appendixchessboardwidth\columnwidth}
        {
            \chessboard[
                showmover=true,
                color=yellow!70,
                setfen= rnbqrbk1/pp3Rp1/2p1p1N1/3p1P1Q/3PnB2/2P5/PP3P1P/6K1 w - - 0 1
            ]
        }
\begin{center}
    \textbf{\href{https://lichess.org/analysis/rnbqrbk1/pp3Rp1/2p1p1N1/3p1P1Q/3PnB2/2P5/PP3P1P/6K1 w - - 0 1}{[Analyse on Lichess]}}

\textbf{Closest FENs - \href{https://lichess.org/analysis/rnbqr1k1/pp3Rp1/2p5/3p2N1/3Pn3/2P4P/PP4P1/R1BQ2K1 w - - 1 17}{[1]}}, \textbf{\href{https://lichess.org/analysis/r1bqr1k1/pp4p1/2p2nN1/3n3Q/3P4/2P5/PP3PPP/R3R1K1 w - - 6 20}{[2]}}, \textbf{\href{https://lichess.org/analysis/r1bq2k1/pp2b1p1/4p1N1/3p1r1Q/3P4/8/PP3PPP/R3R1K1 w - - 1 18}{[3]}}
\end{center}

\adjustbox{max width=\appendixchessboardwidth\columnwidth}
        {
            \chessboard[
                showmover=true,
                color=yellow!70,
                setfen= q2rbr1k/2R3pp/p3Q3/1pb1N3/3RnNP1/PB5P/1P5K/8 w - - 0 1
            ]
        }
\begin{center}
    \textbf{\href{https://lichess.org/analysis/q2rbr1k/2R3pp/p3Q3/1pb1N3/3RnNP1/PB5P/1P5K/8 w - - 0 1}{[Analyse on Lichess]}}

\textbf{Closest FENs - \href{https://lichess.org/analysis/b2r1r1k/2R3pp/p7/1p6/5NQ1/1B5P/Pq6/6K1 w - - 0 27}{[1]}}, \textbf{\href{https://lichess.org/analysis/5rk1/p1R3pp/8/2b5/4nP1P/6P1/P5K1/3R4 w - - 4 41}{[2]}}, \textbf{\href{https://lichess.org/analysis/3br1k1/1R3ppp/8/8/6P1/5Q1P/5PK1/4q3 w - - 16 35}{[3]}}
\end{center}


\vfill\null

\subsection{Interference~\citep{spectacularchess}}

\noindent\textbf{Book example:}

\adjustbox{max width=\appendixchessboardwidth\columnwidth}
        {
            \chessboard[
                showmover=true,
                color=yellow!70,
                setfen= 2b2N2/4p3/6R1/7k/2N5/6K1/8/8 w - - 0 1
            ]
            }
\begin{center}
    \textbf{\href{https://lichess.org/analysis/2b2N2/4p3/6R1/7k/2N5/6K1/8/8 w - - 0 1}{[Analyse on Lichess]}}
\end{center}

\noindent\textbf{Selected puzzles:}

\adjustbox{max width=\appendixchessboardwidth\columnwidth}
        {
            \chessboard[
                showmover=true,
                color=yellow!70,
                setfen= 3br1k1/pbpQ1ppp/2R5/1P2q3/B2p1N2/P3B3/5PPP/1n4K1 w
            ]
            }
\begin{center}
    \textbf{\href{https://lichess.org/analysis/3br1k1/pbpQ1ppp/2R5/1P2q3/B2p1N2/P3B3/5PPP/1n4K1 w - - 0 1}{[Analyse on Lichess]}}

\textbf{Closest FENs - \href{https://lichess.org/analysis/2r1r1k1/p4ppp/1pN5/4q3/2Q5/P3B3/5PPP/1n2R1K1 w - - 2 29}{[1]}}, \textbf{\href{https://lichess.org/analysis/6k1/1p2rppp/2R5/8/r1BPp3/P3P3/5PPP/1n4K1 w - - 0 28}{[2]}}, \textbf{\href{https://lichess.org/analysis/6k1/p4ppp/1p1R4/5p2/3P1P2/P3r3/6PP/6K1 w - - 0 28}{[3]}}
\end{center}

1. Re6! \chesscomments{White interferes with the coordination of Black’s queen and rook.} Rxe6 2. Nxe6 \chesscomments{Black cannot recapture due to back-rank issues.} h6 3. Qxd8+ Kh7 4. Nf8+ Kg8 5. Ng6+ Kh7 6. Nxe5 \chesscomments{White ends the combination with overwhelming material.}

\adjustbox{max width=\appendixchessboardwidth\columnwidth}
        {
            \chessboard[
                showmover=true,
                color=yellow!70,
                setfen= 1Q2R3/1p1r1r1k/1b1P2p1/1q3pB1/8/1P5P/P2Rn1PK/8 b
            ]
        }
\begin{center}
    \textbf{\href{https://lichess.org/analysis/1Q2R3/1p1r1r1k/1b1P2p1/1q3pB1/8/1P5P/P2Rn1PK/8 b - - 0 1}{[Analyse on Lichess]}}

\textbf{Closest FENs - \href{https://lichess.org/analysis/4R3/1p1r1p1k/p2N1Qpp/2q5/8/7P/P5PK/8 b - - 7 54}{[1]}}, \textbf{\href{https://lichess.org/analysis/1Q6/4qp1k/1N2p2p/2p3p1/5P2/1P2P2P/P3n1PK/8 b - - 0 30}{[2]}}, \textbf{\href{https://lichess.org/analysis/8/p5pk/2Q1P2p/6p1/3P4/1P5P/P3n1PK/2b3r1 b - - 0 37}{[3]}}
\end{center}

1... Rd8! \chesscomments{Black sacrifices the rook to interfere with White's pieces.} 2. Bxd8 \chesscomments{If instead 2. Rxd8 Qe5+ is mating quickly and if 2. Qxd8 simply Bxd8} Qxe8 3. Rxe2 Qxd8 \chesscomments{Black remains up a piece and should win comfortably with correct play.}

\subsection{Unprotected Position~\citep{creativechess}}

\noindent\textbf{Book example:}

\adjustbox{max width=\appendixchessboardwidth\columnwidth}
        {
            \chessboard[
                showmover=true,
                color=yellow!70,
                setfen= 7N/K7/p7/k1p5/2P2R1P/1P4n1/P3p3/8 w - - 0 1
            ]
            }
\begin{center}
    \textbf{\href{https://lichess.org/analysis/7N/K7/p7/k1p5/2P2R1P/1P4n1/P3p3/8 w - - 0 1}{[Analyse on Lichess]}}
\end{center}


\newpage
\noindent\textbf{Selected puzzles:}

\adjustbox{max width=\appendixchessboardwidth\columnwidth}
        {
            \chessboard[
                showmover=true,
                color=yellow!70,
                setfen= r3r1k1/pp3p1p/2p3p1/2Ppn1qn/PP6/2NBP2P/2Q2PP1/R1B2RK1 b
            ]
            }
\begin{center}
    \textbf{\href{https://lichess.org/analysis/r3r1k1/pp3p1p/2p3p1/2Ppn1qn/PP6/2NBP2P/2Q2PP1/R1B2RK1 b - - 0 1}{[Analyse on Lichess]}}

\textbf{Closest FENs - \href{https://lichess.org/analysis/r3r1k1/pp5p/2p3p1/2Ppnpq1/1P6/3BPP1P/P2Q2P1/R4RK1 w - - 1 18}{[1]}}, \textbf{\href{https://lichess.org/analysis/r4r1k/bpp3p1/p6p/3p1n1q/PP1P4/3B3P/2Q2PP1/R1B2RK1 w - - 1 22}{[2]}}, \textbf{\href{https://lichess.org/analysis/r1b3k1/pp3p1p/2p3p1/2Ppr1qn/8/2NBP2P/PP3PP1/R2Q1RK1 w - - 2 15}{[3]}}
\end{center}

1... Nf3+ 2. Kh1 Qg3! 3. gxf3 Qxh3+ 4. Kg1 Re5 \chesscomments{Ending the game with a rook lift.} 5. f4 Nxf4! 6. exf4 Qg4+ 7. Kh1 Rh5\#.

\adjustbox{max width=\appendixchessboardwidth\columnwidth}
        {
            \chessboard[
                showmover=true,
                color=yellow!70,
                setfen= 2r2b1k/qp3pp1/p3pN1p/8/1P4Q1/P5R1/5PPP/6K1 w
            ]
        }
\begin{center}
    \textbf{\href{https://lichess.org/analysis/2r2b1k/qp3pp1/p3pN1p/8/1P4Q1/P5R1/5PPP/6K1 w - - 0 1}{[Analyse on Lichess]}}

\textbf{Closest FENs - \href{https://lichess.org/analysis/2r2b1k/1pq2pp1/p3pN1p/7Q/P7/2P1P3/6RP/6K1 w - - 4 28}{[1]}}, \textbf{\href{https://lichess.org/analysis/2r2bk1/Bp3p1p/3p2p1/3R1N2/1P6/P7/5PPP/6K1 b - - 0 23}{[2]}}, \textbf{\href{https://lichess.org/analysis/2r3k1/pb3pp1/1p2p2p/4Q3/1P6/P7/5PPP/6K1 b - - 0 26}{[3]}}
\end{center}

1. Qg5! \chesscomments{Defending c1 and threatening 2. Qxg6+ gxh6 3. Rg8\#. Black tries to hold the position.} g6 2.  Rh3 Kg7 3. Rxh6 Rc1+ 4. Qxc1 \chesscomments{Black is forced to give up material to defend checkmate.}

\adjustbox{max width=\appendixchessboardwidth\columnwidth}
        {
            \chessboard[
                showmover=true,
                color=yellow!70,
                setfen= 6k1/8/qR2pr1p/3pNb2/3Q2p1/2P5/1P4PP/6K1 b
            ]
        }
\begin{center}
    \textbf{\href{https://lichess.org/analysis/6k1/8/qR2pr1p/3pNb2/3Q2p1/2P5/1P4PP/6K1 b - - 0 1}{[Analyse on Lichess]}}

\textbf{Closest FENs - \href{https://lichess.org/analysis/6k1/1p6/4p3/p1N1b3/2P2p2/1P6/P5PP/6K1 b - - 0 27}{[1]}}, \textbf{\href{https://lichess.org/analysis/6k1/8/pR1pr1pp/8/3p4/P6P/1P3PP1/5K2 b - - 0 32}{[2]}}, \textbf{\href{https://lichess.org/analysis/6k1/8/1R5p/p3PN2/3r2p1/8/PP3PPP/6K1 b - - 0 31}{[3]}}
\end{center}

1... Bd3! \chesscomments{Black leaves the queen hanging, but the queen cannot be captured due to the mate threat on f1.} 2. Rb8+ \chesscomments{If 2. h3 g3! 3. Rb8+ Kg7 4. Qg4+ Bg6 5. Qxg3 Qa7+ and Black picks up the rook.} Kg7 3. Qxg4+ Bg6 4. h3 Ba7+ \chesscomments{Black picks up the rook.}

\adjustbox{max width=\appendixchessboardwidth\columnwidth}
        {
            \chessboard[
                showmover=true,
                color=yellow!70,
                setfen= r6k/1bb2Bpp/2p1pq2/p2nR2Q/8/1P6/PBP2PPP/6K1 w
            ]
        }
\begin{center}
    \textbf{\href{https://lichess.org/analysis/r6k/1bb2Bpp/2p1pq2/p2nR2Q/8/1P6/PBP2PPP/6K1 w - - 0 1}{[Analyse on Lichess]}}

\textbf{Closest FENs - \href{https://lichess.org/analysis/6k1/1b2rpp1/p1p1p2p/1pN5/8/P6P/1P3PP1/3R2K1 w - - 0 22}{[1]}}, \textbf{\href{https://lichess.org/analysis/6k1/1b1p1pp1/1p5p/4R3/8/1P1B4/r1P2PPP/6K1 w - - 0 26}{[2]}}, \textbf{\href{https://lichess.org/analysis/r6k/1RQ3pp/p4pq1/4n3/8/7P/PP3PP1/6K1 w - - 3 30}{[3]}}
\end{center}

1. Rxd5! e5 \chesscomments{If Qxb2 2. Qxh7+! Kxh7 3. Rh5\#} 2. Bxe5 Bxe5 3. Rxe5 g6 4. Qg5! \chesscomments{The critical move that makes this variation work.} Qxf7 5. Re7 Qf8 6. Qe5+ Kg8 7. Rxb7 Re8 \chesscomments{White seems in trouble due to the back-rank issues, but there is a beautiful finishing move here.} 8. Rg7+! Qxg7 9. Qe8+ Qf8 10. Qxf8 \chesscomments{After the dust has settled, white remains up 2 pawns and easily wins the pawn endgame.}

\adjustbox{max width=\appendixchessboardwidth\columnwidth}
        {
            \chessboard[
                showmover=true,
                color=yellow!70,
                setfen= q2rr2k/5Qp1/1p2p1B1/pb2N1Pp/3P3P/8/2P4n/2K2R2 w - - 0 1
            ]
        }
\begin{center}
    \textbf{\href{https://lichess.org/analysis/q2rr2k/5Qp1/1p2p1B1/pb2N1Pp/3P3P/8/2P4n/2K2R2 w - - 0 1}{[Analyse on Lichess]}}

\textbf{Closest FENs - \href{https://lichess.org/analysis/3r3k/5Qpp/2p5/pp6/3P4/2P5/2P3qP/2K2R2 w - - 1 30}{[1]}}, \textbf{\href{https://lichess.org/analysis/3r3k/5Qp1/1p2p1Pr/p7/2P5/8/q7/5RK1 w - - 3 36}{[2]}}, \textbf{\href{https://lichess.org/analysis/3r3k/p4Qp1/1p6/6Pp/3q3P/8/PP6/1K2R3 b - - 0 31}{[3]}}
\end{center}

\adjustbox{max width=\appendixchessboardwidth\columnwidth}
        {
            \chessboard[
                showmover=true,
                color=yellow!70,
                setfen= 3Bk1r1/5p2/2qbNQ1p/p1pNpPP1/r1p3p1/6Kn/2P5/8 w - - 0 1
            ]
        }
\begin{center}
    \textbf{\href{https://lichess.org/analysis/3Bk1r1/5p2/2qbNQ1p/p1pNpPP1/r1p3p1/6Kn/2P5/8 w - - 0 1}{[Analyse on Lichess]}}

\textbf{Closest FENs - \href{https://lichess.org/analysis/4k3/5p2/4pP2/1pKpP2p/1p5P/8/2P5/8 w - - 4 33}{[1]}}, \textbf{\href{https://lichess.org/analysis/4k3/5p2/2Np2pp/P1p2n2/8/5K2/1PP5/8 w - - 1 42}{[2]}}, \textbf{\href{https://lichess.org/analysis/4k3/5p2/5Pp1/p1ppPP1p/P1p4P/2P3K1/2P5/8 b - - 0 34}{[3]}}
\end{center}

\adjustbox{max width=\appendixchessboardwidth\columnwidth}
        {
            \chessboard[
                showmover=true,
                color=yellow!70,
                setfen= 2r3k1/p2R1nBp/2r3p1/q2N1p2/1bQ2P2/1P2P3/2R2K2/8 w - - 0 1
            ]
        }
\begin{center}
    \textbf{\href{https://lichess.org/analysis/2r3k1/p2R1nBp/2r3p1/q2N1p2/1bQ2P2/1P2P3/2R2K2/8 w - - 0 1}{[Analyse on Lichess]}}

\textbf{Closest FENs - \href{https://lichess.org/analysis/2r3k1/p4r1p/2N3p1/3n4/1b1P1P2/1P2P3/P1RP2K1/7R w - - 1 37}{[1]}}, \textbf{\href{https://lichess.org/analysis/6k1/p2r1p1p/6p1/3N4/R7/5P2/5K2/8 w - - 5 35}{[2]}}, \textbf{\href{https://lichess.org/analysis/5k2/2R2bp1/1r2pp2/2Np3p/3P3P/2K1P1P1/5P2/8 w - - 1 42}{[3]}}
\end{center}


\subsection{XRay Attack}

\noindent\textbf{Selected puzzles:}

\adjustbox{max width=\appendixchessboardwidth\columnwidth}
        {
            \chessboard[
                showmover=true,
                color=yellow!70,
                setfen= 2kr4/5p1p/2p1rQb1/1p1pN3/1q1P4/pP5P/P2RPP2/K1R5 w
            ]
            }
\begin{center}
    \textbf{\href{https://lichess.org/analysis/2kr4/5p1p/2p1rQb1/1p1pN3/1q1P4/pP5P/P2RPP2/K1R5 w - - 0 1}{[Analyse on Lichess]}}

\textbf{Closest FENs - \href{https://lichess.org/analysis/2rr3k/5pp1/1p2pb1p/pb2N3/3P4/1P5P/PB1R1PP1/1K1R4 w - - 4 27}{[1]}}, \textbf{\href{https://lichess.org/analysis/2k5/3R1p2/p3r3/Qpq1p3/8/2P5/PP3P2/K7 w - - 4 45}{[2]}}, \textbf{\href{https://lichess.org/analysis/2kr4/1b6/p2p2q1/1pp1pN2/2n1P2B/2P2p2/PPQ3P1/1K5R w - - 2 29}{[3]}}
\end{center}

1. Rc6+ \chesscomments{The queen defends the rook via an x-ray.} Rxc6 \chesscomments{The immediate Kb7 transposes to the main line.} 2. Qxd8+! Kb7 3. Qb8+ \chesscomments{White insists on the sacrifice. Black is now forced to accept.} Kxb8 4. Nxc6+ \chesscomments{Winning the queen in the next move. White finishes the variation up a rook.}

\adjustbox{max width=\appendixchessboardwidth\columnwidth}
        {
            \chessboard[
                showmover=true,
                color=yellow!70,
                setfen= 5r1k/pQR3pp/1p1pBb2/1q2nN2/3B2P1/1P1r4/5P1P/6K1 w
            ]
        }
\begin{center}
    \textbf{\href{https://lichess.org/analysis/5r1k/pQR3pp/1p1pBb2/1q2nN2/3B2P1/1P1r4/5P1P/6K1 w - - 0 1}{[Analyse on Lichess]}}

\textbf{Closest FENs - \href{https://lichess.org/analysis/6k1/ppR2ppp/2p1p1b1/8/6P1/1PNr4/P4P2/6K1 w - - 1 25}{[1]}}, \textbf{\href{https://lichess.org/analysis/5r1k/p5p1/1p2Q3/8/3B2Pp/2P4q/5P2/6K1 w - - 4 38}{[2]}}, \textbf{\href{https://lichess.org/analysis/3R1r1k/pp4pp/4n3/1q2R3/8/P5P1/5P1P/6K1 w - - 0 34}{[3]}}
\end{center}

1. Bc4 Nf3+ \chesscomments{If Rd1+ instead, then 2. Kg2 Qb4 3. Rxg7! and White wins quickly.} 2. Kg2 Nh4 3. Nxh4 Qg5 \chesscomments{Black attempts a clever maneuver, rerouting the queen towards a better attacking square hoping to complicate matters.} 4. Rxg7! Qxg7 5. Qxg7 Bxg7 8. Bxg7 \chesscomments{The x-ray attack of White’s bishop results in massive liquidation.} Kxg7 9. Bxd3 \chesscomments{White ends the combination with a bishop and knight for a rook, in a technically winning endgame although it will require careful play.}

\adjustbox{max width=\appendixchessboardwidth\columnwidth}
        {
            \chessboard[
                showmover=true,
                color=yellow!70,
                setfen= r2q1rk1/5ppp/2bp4/p3pPP1/1p2Bb2/5Q2/PPP5/1K1R3R w - - 0 1
            ]
        }
\begin{center}
    \textbf{\href{https://lichess.org/analysis/r2q1rk1/5ppp/2bp4/p3pPP1/1p2Bb2/5Q2/PPP5/1K1R3R w - - 0 1}{[Analyse on Lichess]}}

\textbf{Closest FENs - \href{https://lichess.org/analysis/r2q1rk1/p5pp/1p2bp2/2pp2nP/5Q2/5N1B/PPP2P2/1K1R3R w - - 0 21}{[1]}}, \textbf{\href{https://lichess.org/analysis/r2q1rk1/5ppp/2bP4/p1N1p1b1/1p4P1/3Q1P2/PPP3B1/2KR3R w - - 3 20}{[2]}}, \textbf{\href{https://lichess.org/analysis/r1q2r1k/6pp/p5p1/1p1Q2P1/4Pb2/8/PPP5/1K1R3R w - - 2 26}{[3]}}
\end{center}


\subsection{Paralysis~\citep{spectacularchess}}

\noindent\textbf{Book example:}

\adjustbox{max width=\appendixchessboardwidth\columnwidth}
        {
            \chessboard[
                showmover=true,
                color=yellow!70,
                setfen= 8/1P3kp1/5PN1/6PK/3p4/p2n4/3B3b/8 w - - 0 1
            ]
            }
\begin{center}
    \textbf{\href{https://lichess.org/analysis/8/1P3kp1/5PN1/6PK/3p4/p2n4/3B3b/8 w - - 0 1}{[Analyse on Lichess]}}
\end{center}


\newpage
\noindent\textbf{Selected puzzles:}

\adjustbox{max width=\appendixchessboardwidth\columnwidth}
        {
            \chessboard[
                showmover=true,
                color=yellow!70,
                setfen= 5qrk/5p1p/1R1n1PpR/4p2N/3pP1Q1/1r1P3P/6PK/8 w
            ]
            }
\begin{center}
    \textbf{\href{https://lichess.org/analysis/5qrk/5p1p/1R1n1PpR/4p2N/3pP1Q1/1r1P3P/6PK/8 w - - 0 1}{[Analyse on Lichess]}}

\textbf{Closest FENs - \href{https://lichess.org/analysis/5Qrk/8/q4p1p/4p3/3pP3/3P3P/6PK/8 w - - 2 42}{[1]}}, \textbf{\href{https://lichess.org/analysis/6rk/5p1p/3p1PpQ/p7/1pp2R2/P1q4P/6PK/8 w - - 0 39}{[2]}}, \textbf{\href{https://lichess.org/analysis/6rk/6p1/1R5p/3q4/p4P1Q/7P/6PK/8 w - - 0 40}{[3]}}
\end{center}

1. Ng7 \chesscomments{White sets up 2. Rxh7+ 3. Qh4\#. Black is forced to act quickly.} Rxg7 2. Rxb3! \chesscomments{White ignores the situation on the king side and instead aims to suffocate Black’s pieces.} Ne8 3. Qh4 \chesscomments{Black’s position is completely paralysed. We show a line that highlights the theme.} Qc5 4. Rb8! Qc6 5. Qg5! \chesscomments{Black is never able to save the rook due to mate threats with Rxh7+.} Kg8 6. Qxe5! Kh8 7. Rxe8+ Rg8 8. Rxg8+ Kxg8 9. Qb8+ Qe8 10. Qxe8\#.

\adjustbox{max width=\appendixchessboardwidth\columnwidth}
        {
            \chessboard[
                showmover=true,
                color=yellow!70,
                setfen= 8/6pk/7p/R3n3/8/1BP5/Pr5P/7K b
            ]
        }
\begin{center}
    \textbf{\href{https://lichess.org/analysis/8/6pk/7p/R3n3/8/1BP5/Pr5P/7K b - - 0 1}{[Analyse on Lichess]}}

\textbf{Closest FENs - \href{https://lichess.org/analysis/8/6pk/7p/4pR2/8/2P5/1r4PP/7K b - - 0 32}{[1]}}, \textbf{\href{https://lichess.org/analysis/8/5ppk/3N3p/R7/4P3/bP5P/r4PP1/6K1 b - - 2 29}{[2]}}, \textbf{\href{https://lichess.org/analysis/8/6pk/7p/3p4/4Q3/7P/6P1/6K1 b - - 0 49}{[3]}}
\end{center}

1... Nf3 2. Rh5 g5! \chesscomments{Black aims to trap the White rook.} 3. Bd5 Nh4! \chesscomments{If 3. Rh3 g4! 4. Rh5 Kg6 and White loses the rook and is getting mated quickly.} 4. Be4+ Kg7 5. Kg1 Rxa2 \chesscomments{White’s rook is paralysed and Black should be able to eventually win this position without much resistance.}

\adjustbox{max width=\appendixchessboardwidth\columnwidth}
        {
            \chessboard[
                showmover=true,
                color=yellow!70,
                setfen= 8/3k1p2/3Pb3/2K5/7p/p7/3R2P1/8 b
            ]
        }
\begin{center}
    \textbf{\href{https://lichess.org/analysis/8/3k1p2/3Pb3/2K5/7p/p7/3R2P1/8 b - - 0 1}{[Analyse on Lichess]}}

\textbf{Closest FENs - \href{https://lichess.org/analysis/8/2B5/1Pk5/K7/5ppp/8/5P1P/8 b - - 1 50}{[1]}}, \textbf{\href{https://lichess.org/analysis/8/2k1N3/2P5/3K4/6pp/8/6P1/8 b - - 1 57}{[2]}}, \textbf{\href{https://lichess.org/analysis/8/2k5/2P5/1K6/5ppp/8/5PPP/8 b - - 0 45}{[3]}}
\end{center}

1... a2 2. Rd1 Bf5 \chesscomments{Black threatens Bb1 which would secure the promotion.} 3. Rf1 Kd8! \chesscomments{Black cannot immediately play Bb1?? due to Rxf7+. The slow Kd8 instead wins!} 4. Ra1 Bb1 \chesscomments{White is now completely paralysed. Black has the simple plan of pushing the f and h pawns and wins this position without much effort.}

\vfill\null

\subsection{Bristol~\citep{spectacularchess}}

\noindent\textbf{Book example:}

\adjustbox{max width=\appendixchessboardwidth\columnwidth}
        {
            \chessboard[
                showmover=true,
                color=yellow!70,
                setfen= R7/Pp2b1p1/8/8/4p3/8/k1K3p1/8 w - - 0 1
            ]
            }
\begin{center}
    \textbf{\href{https://lichess.org/analysis/R7/Pp2b1p1/8/8/4p3/8/k1K3p1/8 w - - 0 1}{[Analyse on Lichess]}}
\end{center}

\noindent\textbf{Selected puzzles:}

\adjustbox{max width=\appendixchessboardwidth\columnwidth}
        {
            \chessboard[
                showmover=true,
                color=yellow!70,
                setfen= r1b2k1r/ppp2Bpp/2n2n2/b3N3/3q4/1Q6/P4PPP/RNB2K1R w
            ]
            }
\begin{center}
    \textbf{\href{https://lichess.org/analysis/r1b2k1r/ppp2Bpp/2n2n2/b3N3/3q4/1Q6/P4PPP/RNB2K1R w - - 0 1}{[Analyse on Lichess]}}

\textbf{Closest FENs - \href{https://lichess.org/analysis/r1b2k1r/ppp3pp/5n2/b3N3/3P4/1Q6/PP3PPP/RNq2K1R w - - 0 14}{[1]}}, \textbf{\href{https://lichess.org/analysis/r1b2k1r/ppp2ppp/1bn2n2/1B2N3/3q4/1Q6/PP3PPP/RNB1R1K1 w - - 0 12}{[2]}}, \textbf{\href{https://lichess.org/analysis/r1b2k1r/pppp1Bpp/5n2/b7/1q6/1Q6/PB3PPP/RN3RK1 w - - 2 14}{[3]}}
\end{center}

1. Ba3+ Nb4 \chesscomments{1... Ne7 is again met by 2. Bg8 and 1... Bb4 is followed by simply 2. Nxc6.} 2. Bg8! \chesscomments{The Bristol move, making space for the White queen.} Be7 3. Bb2 \chesscomments{White does not hurry with Bf7+ and first plays an intermezzo to solidify their position.} Qc5 4. Qf7+ Kd8 5. Nc3 \chesscomments{White enjoys a decisive advantage with a safer king and very active pieces.}



\vfill\null

\subsection{King on Tour~\citep{tigerchess}}

\noindent\textbf{Book example:}

\adjustbox{max width=\appendixchessboardwidth\columnwidth}
        {
            \chessboard[
                showmover=true,
                color=yellow!70,
                setfen= 5rk1/4Qpq1/bpp3p1/p6p/4B2R/1P2R1P1/P3PP1P/3r2K1 w - - 0 1
            ]
            }
\begin{center}
    \textbf{\href{https://lichess.org/analysis/5rk1/4Qpq1/bpp3p1/p6p/4B2R/1P2R1P1/P3PP1P/3r2K1 w - - 0 1}{[Analyse on Lichess]}}
\end{center}

\noindent\textbf{Selected puzzles:}


\adjustbox{max width=\appendixchessboardwidth\columnwidth}
        {
            \chessboard[
                showmover=true,
                color=yellow!70,
                setfen= r5k1/bb1p1Np1/pq1P2n1/nP5Q/2B2r2/8/1B3PPP/RN3RK1 b - - 0 1
            ]
            }
\begin{center}
    \textbf{\href{https://lichess.org/analysis/r5k1/bb1p1Np1/pq1P2n1/nP5Q/2B2r2/8/1B3PPP/RN3RK1 b - - 0 1}{[Analyse on Lichess]}}

\textbf{Closest FENs - \href{https://lichess.org/analysis/r5k1/nb1p1pp1/p2P2n1/1p5Q/8/7R/P4PPP/q4BK1 w - - 0 22}{[1]}}, \textbf{\href{https://lichess.org/analysis/r4rk1/b1p2ppp/p1P2q2/P5B1/6b1/5N2/1P3PPP/R2Q1RK1 b - - 2 16}{[2]}}, \textbf{\href{https://lichess.org/analysis/r4rk1/pbp2Npp/1pq2n2/8/2B5/2Q5/PB3PPP/R4RK1 b - - 0 18}{[3]}}
\end{center}

\vfill\null

\adjustbox{max width=\appendixchessboardwidth\columnwidth}
        {
            \chessboard[
                showmover=true,
                color=yellow!70,
                setfen= 1rb4r/3N3k/3pNQ2/p5p1/q1P1P1p1/1R1P1p2/6Pp/6K1 w - - 0 1
            ]
            }
\begin{center}
    \textbf{\href{https://lichess.org/analysis/1rb4r/3N3k/3pNQ2/p5p1/q1P1P1p1/1R1P1p2/6Pp/6K1 w - - 0 1}{[Analyse on Lichess]}}

\textbf{Closest FENs - \href{https://lichess.org/analysis/1r4r1/7k/1p1p3p/p4Rp1/b1P5/3B1P2/6PP/6K1 w - - 0 33}{[1]}}, \textbf{\href{https://lichess.org/analysis/2b5/6q1/7p/6p1/p1P3k1/1P6/6PQ/6K1 w - - 0 48}{[2]}}, \textbf{\href{https://lichess.org/analysis/6k1/6p1/2p4p/5p2/1PP1p3/3q3P/5PP1/1Q4K1 w - - 1 32}{[3]}}
\end{center}

\adjustbox{max width=\appendixchessboardwidth\columnwidth}
        {
            \chessboard[
                showmover=true,
                color=yellow!70,
                setfen= r1b2rk1/qp1ppp2/pbn3pQ/4P1Bp/2PP3N/6n1/1P1R2PK/5B2 w - - 0 1
            ]
            }
\begin{center}
    \textbf{\href{https://lichess.org/analysis/r1b2rk1/qp1ppp2/pbn3pQ/4P1Bp/2PP3N/6n1/1P1R2PK/5B2 w - - 0 1}{[Analyse on Lichess]}}

\textbf{Closest FENs - \href{https://lichess.org/analysis/r1b2rk1/p1p1pp1p/1p4pQ/3P2B1/4N3/6P1/P3RbPK/5q2 w - - 0 24}{[1]}}, \textbf{\href{https://lichess.org/analysis/r5k1/pp2Qppp/2p1bn2/5q2/8/6P1/PP1R2PK/8 w - - 6 29}{[2]}}, \textbf{\href{https://lichess.org/analysis/r1b2rk1/1p3pp1/p1p1n3/6QN/P3q3/8/2P3PK/8 w - - 3 30}{[3]}}
\end{center}

\adjustbox{max width=\appendixchessboardwidth\columnwidth}
        {
            \chessboard[
                showmover=true,
                color=yellow!70,
                setfen= 1qb5/5k2/P1Qp1bNp/2pP1P2/2P1pP1P/8/3rB1R1/4K3 b - - 0 1
            ]
            }
\begin{center}
    \textbf{\href{https://lichess.org/analysis/1qb5/5k2/P1Qp1bNp/2pP1P2/2P1pP1P/8/3rB1R1/4K3 b - - 0 1}{[Analyse on Lichess]}}

\textbf{Closest FENs - \href{https://lichess.org/analysis/8/5k2/2p3pK/1pP1p3/1P2P2P/8/8/8 b - - 6 46}{[1]}}, \textbf{\href{https://lichess.org/analysis/8/5pk1/3p3p/2pP1Pp1/2P1R1P1/6P1/2r5/3K4 b - - 4 53}{[2]}}, \textbf{\href{https://lichess.org/analysis/8/7P/2b1k1K1/1pP1P3/1P1p2P1/8/8/8 b - - 0 49}{[3]}}
\end{center}


\subsection{Switchback~\citep{spectacularchess}}

\noindent\textbf{Book example:}

\adjustbox{max width=\appendixchessboardwidth\columnwidth}
        {
            \chessboard[
                showmover=true,
                color=yellow!70,
                setfen= 8/2p2Pp1/p3pN2/2p1P3/2Pk4/3P4/2PK1Pq1/8 w - - 0 1
            ]
            }
\begin{center}
    \textbf{\href{https://lichess.org/analysis/8/2p2Pp1/p3pN2/2p1P3/2Pk4/3P4/2PK1Pq1/8 w - - 0 1}{[Analyse on Lichess]}}
\end{center}


\noindent\textbf{Selected puzzles:}

\adjustbox{max width=\appendixchessboardwidth\columnwidth}
        {
            \chessboard[
                showmover=true,
                color=yellow!70,
                setfen= 3r2rk/ppp1Qp1n/1q2b1nB/5B2/8/3pR2R/P1P2PPP/6K1 w - - 0 1
            ]
            }
\begin{center}
    \textbf{\href{https://lichess.org/analysis/3r2rk/ppp1Qp1n/1q2b1nB/5B2/8/3pR2R/P1P2PPP/6K1 w - - 0 1}{[Analyse on Lichess]}}

\textbf{Closest FENs - \href{https://lichess.org/analysis/2r2r1k/pp3p1p/4b3/5B2/8/4R2R/2P2PPP/2K5 w - - 1 25}{[1]}}, \textbf{\href{https://lichess.org/analysis/r1b2r1k/ppp2p2/5qnp/5p1Q/8/3BR2R/P1P2PPP/6K1 w - - 4 21}{[2]}}, \textbf{\href{https://lichess.org/analysis/3r2k1/ppp2p1p/4b1pq/8/8/1B3R2/P1P1Q1PP/6K1 w - - 0 27}{[3]}}
\end{center}



\subsection{Uncategorized}

\adjustbox{max width=\appendixchessboardwidth\columnwidth}
        {
            \chessboard[
                showmover=true,
                color=yellow!70,
                setfen= 2R5/3r2pR/p2k1q2/4pBb1/Pp6/1P1p1PQ1/1P4PP/7K b
            ]
            }
\begin{center}
    \textbf{\href{https://lichess.org/analysis/2R5/3r2pR/p2k1q2/4pBb1/Pp6/1P1p1PQ1/1P4PP/7K b - - 0 1}{[Analyse on Lichess]}}

\textbf{Closest FENs - \href{https://lichess.org/analysis/2R5/1p2rk2/p3q3/3p1p1P/3P4/1P4Q1/P4PP1/6K1 b - - 1 40}{[1]}}, \textbf{\href{https://lichess.org/analysis/8/3k2pp/1K6/4pp2/Pp6/1P3P2/2P3PP/8 b - - 1 36}{[2]}}, \textbf{\href{https://lichess.org/analysis/8/6pp/3k4/4p3/1p3N2/5P2/1P2K1PP/8 b - - 0 35}{[3]}}
\end{center}

1... d2 2. Bc2 Qf5! \chesscomments{The only winning idea. Black sacrifices the queen to try to promote the pawn.} 3. Bd1 Qc1 \chesscomments{White’s pieces are not coordinated and there is no good response.}

\adjustbox{max width=\appendixchessboardwidth\columnwidth}
        {
            \chessboard[
                showmover=true,
                color=yellow!70,
                setfen= 8/kb6/1p1p1p2/p1pPpP1p/P1P1P2P/2P4K/3N4/8 w - - 0 1
            ]
            }
\begin{center}
    \textbf{\href{https://lichess.org/analysis/8/kb6/1p1p1p2/p1pPpP1p/P1P1P2P/2P4K/3N4/8 w - - 0 1}{[Analyse on Lichess]}}

\textbf{Closest FENs - \href{https://lichess.org/analysis/8/2k5/3p1p2/2pPpP1p/K1P1P2P/8/8/8 w - - 14 84}{[1]}}, \textbf{\href{https://lichess.org/analysis/8/5b2/1p1p1p1k/pPpPpP1p/P1P1P2K/6P1/3N4/8 w - - 4 46}{[2]}}, \textbf{\href{https://lichess.org/analysis/8/8/1p1p1k2/p1p4p/P1P1P2P/2P2K2/8/8 w - - 2 34}{[3]}}
\end{center}


\section{Puzzles generated with evolutionary search}\label{sec:es_puzzles}

The following puzzles are selected from the ones generated with evolutionary search.

As briefly discussed in the Introduction, this method contrasts sharply with our generative modeling approaches. Instead of learning from data, this method relies on applying random mutations and perturbations to a population of chess positions, and directly optimizes for counter intuitiveness check. The optimization process does not enforce realism constraints, which is a feature of our generative approaches and so the puzzles produced with this method often diverges significantly from expert-level games. The allows for generating beyond what is present within the training data, demonstrating a powerful alternative for creative chess puzzle generation.

\adjustbox{max width=\appendixchessboardwidth\columnwidth}{
    \chessboard[
        showmover=true,
        color=yellow!70,
        setfen= 2rRnNk1/Qb2q3/p1P3p1/3p3p/1pr5/P1p1P1qB/3p1PPP/3Q1RK1 b
    ]
    }
\begin{center}
    \textbf{\href{https://lichess.org/analysis/2rRnNk1/Qb2q3/p1P3p1/3p3p/1pr5/P1p1P1qB/3p1PPP/3Q1RK1 b - - 0 1}{[Analyse on Lichess]}}
\end{center}

1... Qxh3 \chesscomments{Giving up the queen for the bishop is the right decision here. The other moves fail tactically. For example 1...Qc7 2. Rd7 or 1... Qb8 2. Qxb8 or 1... Qg5 2. cxb7 Rxd8 3. Ne6 Qf6 4. Nxd8 Rc7 5. Qa8 is winning for White} 2. gxh3 Rxc6 3. Rxc8 Qg5+ 4. Kh1 Bxc8 5. Qf3 Qf5 6. Qxf5 Bxf5 7. Qd4 Kxf8 8. Qxb4 Nd6 \chesscomments{And the pawns on the queenside can’t be stopped.}

\adjustbox{max width=\appendixchessboardwidth\columnwidth}{
    \chessboard[
        showmover=true,
        color=yellow!70,
        setfen= 7K/6P1/5n1R/2N5/8/8/2q1k3/8 b
    ]
    }
\begin{center}
    \textbf{\href{https://lichess.org/analysis/7K/6P1/5n1R/2N5/8/8/2q1k3/8 b - - 0 1}{[Analyse on Lichess]}}
\end{center}

1... Qc4! \chesscomments{Not 1... Qxc5 2. Rxf6 Qh5+ 3. Kg8 and while Black can hold the draw, there is no path to a win here. Other tries don’t work either, for example 1... Qf5 2. Nd7! Qxd7 3. Rxf6 Qh3+ 4. Kg8 Qc8+ 5. Rf8 and this is a draw. Another try would be 1... Qc3, but then White can just play Ne6 and a draw would ensure shortly.} 2. Rxf6 Qh4+ 3. Kg8 Qxf6 4. Nd7 Qf5 5. Nf8 Ke3 \chesscomments{and Black can start bringing the king forward. If White tries 6. Kh8 Qe5 7. Kg8 Qh5 8. Ne6 Ke4 and the king keeps approaching.}

\adjustbox{max width=\appendixchessboardwidth\columnwidth}{
    \chessboard[
        showmover=true,
        color=yellow!70,
        setfen= 5Q2/8/5K2/6N1/k1P1B3/1qQ2N2/8/8 b
    ]
    }
\begin{center}
    \textbf{\href{https://lichess.org/analysis/5Q2/8/5K2/6N1/k1P1B3/1qQ2N2/8/8 b - - 0 1}{[Analyse on Lichess]}}
\end{center}

\chesscomments{Despite White’s overwhelming material advantage, Black manages to salvage a draw.} 1... Qb6+ 2. Ke5 Qf6+ 3. Kd5 Qd6+ 4. Kxd6.

 \adjustbox{max width=\appendixchessboardwidth\columnwidth}{
    \chessboard[
        showmover=true,
        color=yellow!70,
        setfen= 4B3/1p1pK3/pP6/P1P1k1n1/2P3B1/p6n/7P/8 w
    ]
    }
\begin{center}
    \textbf{\href{https://lichess.org/analysis/4B3/1p1pK3/pP6/P1P1k1n1/2P3B1/p6n/7P/8 w - - 0 1}{[Analyse on Lichess]}}
\end{center}

\chesscomments{Among the 3 possible captures on d7 (and other alternatives) only one is correct.} 1. Bgxd7 a2 2. c6 a1=Q 3. cxb7 Qa3+ 4. Kd8 Ne6+ 5. Kc8 Qc5+ 6. Bc6 Qe7 7. b8=Q+ \chesscomments{wins for White - as an example line that follows. If White were to instead play 1. Kxd7, then Black would lose if following up with 1... a2, but wins in case of 1. Kxd7 Ne4 2. c6 Nc5+ 3. Kc7 bxc6 4. Kxc6 Kd4 5. Bxh3 a2 6. b7 Nxb7 7. Kxb7 a1=Q. It’s important to note the following variation: 1. Bgxd7 Ne4 2.. c6 Nd6 3. c5, which is why Ne4 can’t save Black in the solution. Finally, in case of the other bishop capture: 1. Bexd7 a2 2. c6 a1=Q and now 3. cxb7 wouldn’t work (in contrast to the solution) 3... Qa3+ 4. Kd8 Nf7+ 52. Kc7 Qd6+ as an example line - the knight check on f7 is possible as the bishop is no longer on e8 - other moves don’t work either (c7 or alternative king moves).}

\adjustbox{max width=\appendixchessboardwidth\columnwidth}{
    \chessboard[
        showmover=true,
        color=yellow!70,
        setfen= R1qrr2k/1p1bn2p/q1pb3q/5pQN/3P3N/4Q3/B4PPP/nB1R2K1 b
    ]
    }
\begin{center}
\textbf{\href{https://lichess.org/analysis/R1qrr2k/1p1bn2p/q1pb3q/5pQN/3P3N/4Q3/B4PPP/nB1R2K1 b - - 0 1}{[Analyse on Lichess]}}
\end{center}

1... Qe2! \chesscomments{Moving the queen from one square where it can be captured to another.} 2. Qxe2 Qxg5 3. Rxc8 Qxh4 4. g3 Nxc8 \chesscomments{and Black is winning.}

\adjustbox{max width=\appendixchessboardwidth\columnwidth}{
    \chessboard[
        showmover=true,
        color=yellow!70,
        setfen= 8/2k1p3/4p3/3pp1pp/2pPPPP1/5p1B/P4K2/8 w
    ]
    }
\begin{center}
\textbf{\href{https://lichess.org/analysis/8/2k1p3/4p3/3pp1pp/2pPPPP1/5p1B/P4K2/8 w - - 0 1}{[Analyse on Lichess]}}
\end{center}

\chesscomments{In this position, there are many moves to consider, and precise calculation is needed to establish that only one of them wins for White.} 1. Kxf3 hxg4+ 2. Bxg4 dxe4+ 3. Kxe4 c3 4. Kd3 exf4 5. Bxe6 \chesscomments{Let’s consider the alternatives. 1. dxe5 dxe4 2. fxg5 c3 3. Ke3 f2 4. Bf1 h4 5. g6 c2 6. Kd2 h3 7. Bxh3 e3+ 8. Kxc2 e2 9. g7 e1=Q 10. g8=Q Qxe5 is a draw instead. 1. gxh5 c3 2. h6 c2 3. h7 c1=Q 4. h8=Q Qd2+ 5. Kxf3 dxe4+ is winning for Black, because 6. Kxe4 is impossible due to the threat of 6... Qe2\# For the same reason 1. fxg5 fails. Finally, 1. exd5 c3 2. Ke3 exf4+ 3. Kd3 hxg4 4. Bxg4 f2 5. Be2 g4 is winning for Black.}

\adjustbox{max width=\appendixchessboardwidth\columnwidth}{
    \chessboard[
        showmover=true,
        color=yellow!70,
        setfen= 3N4/2k5/8/3ppppp/P2PPPP1/8/b2P1K2/8 b
    ]
    }
\begin{center}
    \textbf{\href{https://lichess.org/analysis/3N4/2k5/8/3ppppp/P2PPPP1/8/b2P1K2/8 b - - 0 1}{[Analyse on Lichess]}}
\end{center}

\chesscomments{Of all the possible pawn captures, only one is winning for Black}. 1... fxg4 2. Ne6+ Kd6 3. Nxg5 exf4 4. e5+ Ke7 5. a5 Bc4 \chesscomments{and White can’t stop the Black pawns. Interestingly, capturing the hanging White knight on the first move would have lost the game for Black. 1... Kxd8 2. gxh5 Ke7 3. exd5 exf4 4. h6 Kf6 5. h7 Kg7 6. d6 Be6 7. a5 Bc8 8. a6.}

\vfill\null

\adjustbox{max width=\appendixchessboardwidth\columnwidth}{
    \chessboard[
        showmover=true,
        color=yellow!70,
        setfen= 2B2b1k/1pP5/1K4PP/pB5p/1R1r4/Pb2pppb/1P3B2/8 b
    ]
    }
\begin{center}
\textbf{\href{https://lichess.org/analysis/2B2b1k/1pP5/1K4PP/pB5p/1R1r4/Pb2pppb/1P3B2/8 b- - 0 1}{[Analyse on Lichess]}}
\end{center}

\chesscomments{In what is a very messy position, there is a narrow path to a win.} 1... Rd6+ (1... axb4 2. Bxh3 Rd6+ 3. Ka7 Bxh6 4. c8=Q+ Bg8 5. Bxg3 Rxg6 6. Be5+ \chesscomments{would be winning for White instead. The other captures on move 1 fail as well.}) 2. Ka7 (2. Kxb7 Bd5+ 3. Ka7 Bxc8) Bxc8 3. Bxg3 Rxg6 4. Be5+ Kh7 5. Rxb3 Bc5+ 6. Kb8 Rg8 7. Rd3 Bb6 8. Rd8 Rxd8 9. cxd8=Q Bxd8 10. Kxc8 Kxh6 11. Kxd8 Kg5.

\adjustbox{max width=\appendixchessboardwidth\columnwidth}{
    \chessboard[
        showmover=true,
        color=yellow!70,
        setfen= 8/PP2qk1P/ppp3r1/5p1n/4r3/4NP2/5KRP/2rR4 w
    ]
    }
\begin{center}
    \textbf{\href{https://lichess.org/analysis/8/PP2qk1P/ppp3r1/5p1n/4r3/4NP2/5KRP/2rR4 w - - 0 1}{[Analyse on Lichess]}}
\end{center}

\chesscomments{In what is aesthetically a very pleasing position, there are three different pawns that can be promoted on the next move, and multiple additional promising captures to calculate as well - for example, the hanging Black rook on c1. However, it turns out that there is only one winning move, and it is an under-promotion!} 1. h8=N+ Kg7 2. Rxg6+ Kh7 3. b8=Q Rc2+ 4. Kf1.

\adjustbox{max width=\appendixchessboardwidth\columnwidth}{
    \chessboard[
        showmover=true,
        color=yellow!70,
        setfen= 1R3N2/P4P1p/2P3nk/1p6/pb2B3/2P3r1/1r4PK/1Q1B1r2 b
    ]
    }
\begin{center}
    \textbf{\href{https://lichess.org/analysis/8/1R3N2/P4P1p/2P3nk/1p6/pb2B3/2P3r1/1r4PK/1Q1B1r2 b - - 0 1}{[Analyse on Lichess]}}
\end{center}

\chesscomments{Despite the White queen hanging on b1, Black needs to find a different motif to win in this position.} 1... Bd6 2. Qxb2 Rgf3+ 3. g3 Bxg3+ 4. Kh3 Rh1+ 5. Kg2 Nh4+ 6. Kxh1 Rf1\#.

\adjustbox{max width=\appendixchessboardwidth\columnwidth}{
    \chessboard[
        showmover=true,
        color=yellow!70,
        setfen= N2r2k1/2qbPp2/p6P/1pB1p2b/2p3n1/P1P1prNp/BPbQ2PN/R4RK1 b
    ]
    }
\begin{center}
    \textbf{\href{https://lichess.org/analysis/N2r2k1/2qbPp2/p6P/1pB1p2b/2p3n1/P1P1prNp/BPbQ2PN/R4RK1 b - - 0 1}{[Analyse on Lichess]}}
\end{center}

\chesscomments{In this astoundingly chaotic and complicated position with many possible captures, there is only one move that secures the win. Perhaps more interestingly, interjecting a check on f1 would throw away the advantage, which is rarely the case! This is especially puzzling at the first glance since the rook is otherwise hanging on f3. Yet, there are specific tactical reasons for why this resource is not available. The correct capture is the capture on a8, one example line being: } 1... Rxa8 2. Qd5 Rxg3 3. Nxg4 Bc6 4. Nf6+ Kh8 5. Nxh5 Bxd5 6. Nxg3 Qxc5 \chesscomments{where Black is winning. Obviously this line is not forced, but the alternatives don’t affect the outcome. So, let’s look at why Rxf1 fails specifically on the first move of the problem: 1... Rxf1+ 2. Rxf1 Rxa8 3. Bxe3 Nxe3 4. Qxe3 Qc6 5. Qg5+ Bhg6 6. Nh5 would be winning for White. In contrast, with the rook still there, Black is able to generate simultaneous threats against e3 and g2 in the same variation, under the main line of the solutions: 1... Rxa8 2. Bxe3 Rxg3 3. Nxg4 Bdxg4 4. Qxc2 Rxg2+ 5. Qxg2 hxg2.}

\adjustbox{max width=\appendixchessboardwidth\columnwidth}{
    \chessboard[
        showmover=true,
        color=yellow!70,
        setfen= 3k1r2/1ppb2pQ/p1nq1nPp/2b1Np2/2B3b1/2QPPp2/P1r3PN/R1BQ1RK1 b
    ]
    }
\begin{center}
\textbf{\href{https://lichess.org/analysis/3k1r2/1ppb2pQ/p1nq1nPp/2b1Np2/2B3b1/2QPPp2/P1r3PN/R1BQ1RK1 b - - 0 1}{[Analyse on Lichess]}}
\end{center}

\chesscomments{It is possible to capture either of the two White queens on c3 and h7 respectively, but neither is the best move - the best move involves ignoring the opportunity and giving a check instead.} 1... Rxg2+ 2. Kh1 Nxe5 3. Qxg7 Nxg6 4. Nxf3 Qg3 \chesscomments{is the winning recipe. If Black were to have been tempted by material instead: 1... Rxc3 2. Nxd7 Qxd7 3. gxf3 Bh5 4. Bb2 Nxh7 5. gxh7 Rxc4 6. dxc4 Rh8 7. Bxg7 Rxh7 8. Qxd7+ Kxd7 9. Bd4 Nxd4 10. Rfd1 Bd6 11. Rxd4 Kc6 12. f4 Bc5 13. Rd3 Be2 14. Rc3 is only a draw - as an example line that may follow from the capture.}
\vfill\null

\adjustbox{max width=\appendixchessboardwidth\columnwidth}{
    \chessboard[
        showmover=true,
        color=yellow!70,
        setfen= r4Nk1/2p2pb1/p1n3Bq/1pB1p3/1r1NPpn1/P1pB1N2/RP1Q1P1P/5RK1 b
    ]
    }
\begin{center}
    \textbf{\href{https://lichess.org/analysis/r4Nk1/2p2pb1/p1n3Bq/1pB1p3/1r1NPpn1/P1pB1N2/RP1Q1P1P/5RK1 b - - 0 1}{[Analyse on Lichess]}}
\end{center}

\chesscomments{Black needs to find the correct capture, among the available options.} 1... Nxd4 2. Bxd4 cxd2 3. axb4 exd4 4. Bf5 Qh3 \chesscomments{is winning for Black, unlike the alternatives. If instead: 1... Rxd4 2. bxc3 Rxd3 3. Qxd3 Bxf8 4. Bxf7+ Kxf7 5. Qd5+ Kg7 6. Qd7+ Kh8 7. Qxg4 or 1... cxd2 2. Nf5 Qh3 3. Bxf7+ Kxf7 4. Ng5+ or another example line (with possible deviations at several points): 1... exd4 2. bxc3 fxg6 3. Ne6 Nce5 4. Neg5 dxc3 5. Qd1 Bf6 6. Bxb4 Bxg5 7. h4 Bxh4 8. Bxc3.}

\adjustbox{max width=\appendixchessboardwidth\columnwidth}{
    \chessboard[
        showmover=true,
        color=yellow!70,
        setfen= B2NRnk1/pp3pBp/1Pp2n2/p3B3/Pq6/PrN4P/1PpQ2K1/3r4 w
    ]
    }
\begin{center}
    \textbf{\href{https://lichess.org/analysis/B2NRnk1/pp3pBp/1Pp2n2/p3B3/Pq6/PrN4P/1PpQ2K1/3r4 w - - 0 1}{[Analyse on Lichess]}}
\end{center}

\chesscomments{Of several possible first moves, only one wins. One possible continuation would be: } 1. Qf2 c1=Q 2. axb4 Qg5+ 3. Bg3 Kxg7 4. Nxd1 Qd5+ 5. Kh2 Nxe8 6. bxa7 Qxd8 7. Bxb7 Nc7 8. Ne3 Kg8 9. Ng4 Rxg3 10. Qxg3 Nfe6 11. Nh6+ Kf8 12. Qg8+ Ke7 13. Qxf7+ Kd6 \chesscomments{Moving the queen to e2 instead fails due to: 1. Qe2 Rg1+ 2. Kxg1 c1=Q+ and if 1. Qg5 c1=Q.}

\adjustbox{max width=\appendixchessboardwidth\columnwidth}{
    \chessboard[
        showmover=true,
        color=yellow!70,
        setfen= 2R1b1Q1/pp1Ppp2/1b1rnN1P/4K3/pP1P2PP/2P2N1P/1Pn1kbn1/8 w
    ]
    }
\begin{center}
    \textbf{\href{https://lichess.org/analysis/2R1b1Q1/pp1Ppp2/1b1rnN1P/4K3/pP1P2PP/2P2N1P/1Pn1kbn1/8 w - - 0 1}{[Analyse on Lichess]}}
\end{center}

\chesscomments{To win, amidst the chaos, White needs to play a quiet king move in the centre of the board.} 1. Ke4 Bxd7 2. Nxd7 \chesscomments{If instead 1. Qh7 Bg3+ 2. Ke4 Nc5+ 3. Rxc5 Re6+ 4. Re5 Rxf6 5. Ng1+ Kd2 6. Nf3+ Ke2 7. Ng1+ the threats against the White king secure Black a draw.}

\adjustbox{max width=\appendixchessboardwidth\columnwidth}{
    \chessboard[
        showmover=true,
        color=yellow!70,
        setfen= 1Brbr1k1/1pP1N1pp/3bqbBQ/R2p4/R3b3/P1R1p3/1P4PP/1B1nbRK1 b
    ]
    }
\begin{center}
    \textbf{\href{https://lichess.org/analysis/1Brbr1k1/1pP1N1pp/3bqbBQ/R2p4/R3b3/P1R1p3/1P4PP/1B1nbRK1 b - - 0 1}{[Analyse on Lichess]}}
\end{center}

\chesscomments{Black has 5 different ways of capturing the knight on e7 that is giving check. However, the only correct decision is not to capture it at all!} 1... Kf8 2. Qxh7 Bf2+ 3. Kh1 Kxe7 4. cxd8=Q+ Rexd8 5. Bxd6+ Rxd6 6. Rxc8 e2 7. Re8+ Kd7 8. Bd3 Bxd3 9. Bxd3 exf1=Q+ 10. Bxf1 Qxe8 11. Qd3 Ne3 12. Qb5+ Ke7 13. Qxb7+ Qd7 \chesscomments{seems to be holding on, in an example continuation. Alternatively, 1... B8xe7 2. Qxh7+ Kf8 3. Rxe4 dxe4 4. Qh8+ Qg8 5. Qxg8+ Kxg8 6. Ba2+ Kf8 7. Rh5 is winning for White. Rxe4 comes up as a motif in some of the other lines, for alternative first-move captures.}

\adjustbox{max width=\appendixchessboardwidth\columnwidth}{
    \chessboard[
        showmover=true,
        color=yellow!70,
        setfen= 8/4p3/p4p2/P1K2b2/BP1p2k1/2P1n3/3P2P1/8 w
    ]
    }
\begin{center}
    \textbf{\href{https://lichess.org/analysis/8/4p3/p4p2/P1K2b2/BP1p2k1/2P1n3/3P2P1/8 w - - 0 1}{[Analyse on Lichess]}}
\end{center}

1. cxd4 Nxg2 2. Kb6 Bc8 3. Bc6 Nf4 4. Bb7 Bxb7 5. Kxb7 Nd5 6. b5 Nb4 7. bxa6 Nxa6 8. Kxa6 f5 9. Kb6 f4 10. a6 f3 11. a7 f2 12. a8=Q f1=Q 13. Qg8+ \chesscomments{and White will pick up the e7 pawn and have a winning position. Alternative first-move captures don’t work for White, for example: 1. Kxd4 Nxg2 2. Kc5 Nf4 3. b5 Nd3+ 4. Kb6 Nb2 5. Bb3 axb5 6. a6 Nc4+ 7. Kxb5 Nd6+ 8. Kc5 Nc8 9. Bd5 Bd3 10. Bb7 Bxa6 11. Bxa6 Nd6 is a draw, and capturing the knight instead loses: 1. dxe3 dxc3 2. Kb6 Bd3.}

\adjustbox{max width=\appendixchessboardwidth\columnwidth}{
    \chessboard[
        showmover=true,
        color=yellow!70,
        setfen= r3r1k1/b4p2/p7/3p1Nqp/4n3/1P2P2B/1BR1pPPP/3QbRKn w
    ]
    }
\begin{center}
    \textbf{\href{https://lichess.org/analysis/r3r1k1/b4p2/p7/3p1Nqp/4n3/1P2P2B/1BR1pPPP/3QbRKn w - - 0 1}{[Analyse on Lichess]}}
\end{center}

1. Rxe2 Nhxf2 2. Qxe1 Nd3 3. Qd1 Nxb2 4. Rxb2 Bxe3+ 5. Nxe3 Nc3 6. Qf3 Qxe3+ 7. Rbf2 Ra7 8. Qxe3 Rxe3 9. Rf5 \chesscomments{With equality. This is ultimately a choice between two possible captures on e2 - with interesting ideas in both lines. If (the wrong move) 1.Qxe2 then after 1...Nhxf2 2. Qxe1 Nxh3+ 3.Kh1, Black plays the beautiful move 3...Nf4! where after 4.Rxf4 Qxf4! (Another nice sacrifice) 5. exf4 Nf2+ 6. Qxf2 Bxf2 Black is suddenly up material. But the reason why Rxe2 works instead is even more subtle. So, let’s look at that same line, just with the rook on e2. 1.Rxe2 Nhxf2 2. Qxe1 Nxh3+ 3. Kh1 Nf4 4. Rxf4 Qxf4 and now instead of exf4, White has a surprising option 5. Ne7+! Rxe7 6. exf4 Nxf2+ 7. Qxf2 Bxf2 8. Rxe7 - because the rook on e7, where it captured the knight, is not defended by the rook on a8, White doesn't end up down an exchange.}

\adjustbox{max width=\appendixchessboardwidth\columnwidth}{
    \chessboard[
        showmover=true,
        color=yellow!70,
        setfen= 2rk1K2/5p1P/2P3Rp/1PRrP3/3p2p1/1p5R/P6R/1r2q3 w
    ]
    }
\begin{center}
    \textbf{\href{https://lichess.org/analysis/2rk1K2/5p1P/2P3Rp/1PRrP3/3p2p1/1p5R/P6R/1r2q3 w - - 0 1}{[Analyse on Lichess]}}
\end{center}

\chesscomments{The Black rook on d5 is hanging with check, but capturing it is, surprisingly, not best.} 1. b6 Qb4 2. Rd6+ Rxd6 3. h8=Q Qxb6 4. exd6 \chesscomments{With mate to follow. If instead 1. Rxd5+ Kc7+ 2. Kxf7 Qf1+ 3. Rf6 Qc4 4. Rd6 gxh3 5. Ke7 Rg1 it would be Black who is winning.}

\adjustbox{max width=\appendixchessboardwidth\columnwidth}{
    \chessboard[
        showmover=true,
        color=yellow!70,
        setfen= 1R2Q1r1/P2nP2r/bP3QnB/R1R3P1/3b3r/6K1/1k3pbR/5b2 b
    ]
    }
\begin{center}
    \textbf{\href{https://lichess.org/analysis/1R2Q1r1/P2nP2r/bP3QnB/R1R3P1/3b3r/6K1/1k3pbR/5b2 b - - 0 1}{[Analyse on Lichess]}}
\end{center}

\chesscomments{There are many possible moves to consider in this chaotic and highly unrealistic board setup. Yet, there is only one solution, and it involves a temporary forced rook sacrifice!} 1... Rg4+ 2. Kxg4 Nxf6+ \chesscomments{And now recapturing doesn’t work because 3. gxf6 Bfe2+ 4. Kg5 Nf8+ 5. Bg7 Be3+ 6. Kf5 f1=Q+ 7. Ke5 Qf4\#} 3. Kg3 Ne4+ 4. Kg4 Bfe2+ 5. Kf5 f1=Q+ 6. Ke6 Nxc5+ 7. Rxc5 Bg4+ 8. Rf5 Qxf5+ 9. Kd6 Qe5\#.


\section{Puzzles adversarial to chess engines}\label{sec:anti_stockfish_puzzles}
The following puzzles were generated with reinforcement learning. These positions are computationally demanding, and often requires significant search-time for Stockfish to determine the optimal line. This is likely because of many pieces on the board, which causes Stockfish to evaluate a number of lines from a large search tree.

\adjustbox{max width=\appendixchessboardwidth\columnwidth}{
    \chessboard[
        showmover=true,
        color=yellow!70,
        setfen= k6r/1pRq1nbq/1B3Qbq/p2p3p/5P2/RP1p2PP/P2p1K2/3B4 w - - 0 1
    ]
    }
\begin{center}
    \textbf{\href{https://lichess.org/analysis/k6r/1pRq1nbq/1B3Qbq/p2p3p/5P2/RP1p2PP/P2p1K2/3B4 w - - 0 1}{[Analyse on Lichess]}}

textbf{Closest FENs - \href{https://lichess.org/analysis/8/p5k1/1q2Qpp1/2p4p/5P2/1P4PP/P4K2/8 w - - 1 34}{[1]}}, \textbf{\href{https://lichess.org/analysis/6k1/p1q2np1/4Qb1p/1p1B4/4PP2/BP1p3P/P7/6K1 w - - 0 32}{[2]}}, \textbf{\href{https://lichess.org/analysis/8/p7/3k2p1/3b3p/5PP1/1P1p4/P4K2/3B4 w - - 0 38}{[3]}}
\end{center}

\adjustbox{max width=\appendixchessboardwidth\columnwidth}{
    \chessboard[
        showmover=true,
        color=yellow!70,
        setfen= q1r2rk1/1r1pRp2/1pp2Bp1/pn5p/4Q3/5P2/1K6/8 w - - 0 1
    ]
    }
\begin{center}
    \textbf{\href{https://lichess.org/analysis/q1r2rk1/1r1pRp2/1pp2Bp1/pn5p/4Q3/5P2/1K6/8 w - - 0 1}{[Analyse on Lichess]}}

\textbf{Closest FENs - \href{https://lichess.org/analysis/3Rrk2/2pR1p2/1pq1pBp1/p3P2p/8/4P1K1/8/8 w - - 0 37}{[1]}}, \textbf{\href{https://lichess.org/analysis/6k1/5p2/5Bp1/p1R4p/8/5r2/1K5P/8 w - - 2 40}{[2]}}, \textbf{\href{https://lichess.org/analysis/6k1/2pr1p2/1p3Bp1/p6p/8/P7/r7/4R1K1 w - - 0 35}{[3]}}
\end{center}

\adjustbox{max width=\appendixchessboardwidth\columnwidth}{
    \chessboard[
        showmover=true,
        color=yellow!70,
        setfen= krNrqb2/1pRp1p2/pP2p2q/N2NP1pp/3n2b1/Q2B4/P4PP1/4K3 w - - 0 1
    ]
    }
\begin{center}
    \textbf{\href{https://lichess.org/analysis/krNrqb2/1pRp1p2/pP2p2q/N2NP1pp/3n2b1/Q2B4/P4PP1/4K3 w - - 0 1}{[Analyse on Lichess]}}

\textbf{Closest FENs - \href{https://lichess.org/analysis/1r1n1r1k/q5pp/1p1P1p2/p3N3/P1N5/1Q2R3/1P4PP/5K2 w - - 1 28}{[1]}}, \textbf{\href{https://lichess.org/analysis/k1qrr3/2p3p1/1pP2p2/p3p2p/Q7/R7/P4PPP/1R4K1 w - - 0 32}{[2]}}, \textbf{\href{https://lichess.org/analysis/k2r4/p1p1qp2/Pp6/1Q4pp/8/2P5/1P3PPP/5K2 w - - 0 26}{[3]}}
\end{center}

\adjustbox{max width=\appendixchessboardwidth\columnwidth}{
    \chessboard[
        showmover=true,
        color=yellow!70,
        setfen= krqn2q1/1bRR2b1/1B2n1p1/3pr1p1/PN1p1NP1/4P3/3Q1P1P/1b4K1 w - - 0 1
    ]
    }
\begin{center}
    \textbf{\href{https://lichess.org/analysis/krqn2q1/1bRR2b1/1B2n1p1/3pr1p1/PN1p1NP1/4P3/3Q1P1P/1b4K1 w - - 0 1}{[Analyse on Lichess]}}

\textbf{Closest FENs - \href{https://lichess.org/analysis/1k1n4/1P6/1K3p2/2p3p1/1N1p2P1/8/5P2/8 w - - 0 48}{[1]}}, \textbf{\href{https://lichess.org/analysis/rr4k1/3R2b1/pB1Rnpp1/P3p2p/1p2P1N1/6P1/5P1P/6K1 w - - 0 39}{[2]}}, \textbf{\href{https://lichess.org/analysis/4r3/RR2nk1p/4qpp1/3p4/PB6/4P3/5P1P/6K1 w - - 10 28}{[3]}}
\end{center}

\newpage
\adjustbox{max width=\appendixchessboardwidth\columnwidth}{
    \chessboard[
        showmover=true,
        color=yellow!70,
        setfen= 2q2bnk/2rb1RpN/p5Q1/npp1pN1P/3p3P/1b4B1/P2K4/2r5 w - - 0 1
    ]
    }
\begin{center}
    \textbf{\href{https://lichess.org/analysis/2q2bnk/2rb1RpN/p5Q1/npp1pN1P/3p3P/1b4B1/P2K4/2r5 w - - 0 1}{[Analyse on Lichess]}}

\textbf{Closest FENs - \href{https://lichess.org/analysis/2q2r1k/4R1pp/4R3/1pp1Qr1P/3p4/6PK/P7/8 w - - 5 40}{[1]}}, \textbf{\href{https://lichess.org/analysis/7k/8/p6p/1p2N3/2p2r1P/5B2/PP2K3/8 w - - 0 39}{[2]}}, \textbf{\href{https://lichess.org/analysis/6nk/2p2R2/p5p1/1p2N1P1/3q4/7P/6K1/8 w - - 0 39}{[3]}}
\end{center}

\adjustbox{max width=\appendixchessboardwidth\columnwidth}{
    \chessboard[
        showmover=true,
        color=yellow!70,
        setfen= r1q2rk1/pp1pRp2/2bP1Bp1/2p4p/7P/3Q4/P2K4/8 w - - 0 1
    ]
    }
\begin{center}
    \textbf{\href{https://lichess.org/analysis/r1q2rk1/pp1pRp2/2bP1Bp1/2p4p/7P/3Q4/P2K4/8 w - - 0 1}{[Analyse on Lichess]}}

\textbf{Closest FENs - \href{https://lichess.org/analysis/r1q2rk1/pp2Rp2/5Qp1/2p4p/2P2P2/P3PKP1/1P6/3R4 b - - 3 32}{[1]}}, \textbf{\href{https://lichess.org/analysis/6k1/pr3p2/3P1Bp1/2p4p/7P/3q4/PP4P1/K3R3 w - - 0 26}{[2]}}, \textbf{\href{https://lichess.org/analysis/r1b3k1/pp1qRp2/3p1Qp1/2pP3p/2P4P/3P4/P5rK/5R2 w - - 0 26}{[3]}}
\end{center}

\newpage
\adjustbox{max width=\appendixchessboardwidth\columnwidth}{
    \chessboard[
        showmover=true,
        color=yellow!70,
        setfen= 2r1q3/N1P1r2k/QNb3p1/B1p4p/RPPnp3/R6P/6PK/8 b - - 0 1
    ]
    }
\begin{center}
    \textbf{\href{https://lichess.org/analysis/2r1q3/N1P1r2k/QNb3p1/B1p4p/RPPnp3/R6P/6PK/8 b - - 0 1}{[Analyse on Lichess]}}

\textbf{Closest FENs - \href{https://lichess.org/analysis/8/4k3/Qpbr4/2p4p/P2p4/1P3P1P/6PK/8 b - - 0 38}{[1]}}, \textbf{\href{https://lichess.org/analysis/8/6pk/5p1p/1Q6/PP1p4/2q4P/6PK/8 b - - 0 41}{[2]}}, \textbf{\href{https://lichess.org/analysis/8/6pk/p6p/1Q6/1P2p3/7P/6PK/8 b - - 0 56}{[3]}}
\end{center}

\adjustbox{max width=\appendixchessboardwidth\columnwidth}{
    \chessboard[
        showmover=true,
        color=yellow!70,
        setfen= 1kqr2r1/R1p1np1p/2p1b2b/1PPpBB1q/P2Q1p1p/4PN2/1RK2P1P/8 w - - 0 1
    ]
    }
\begin{center}
    \textbf{\href{https://lichess.org/analysis/1kqr2r1/R1p1np1p/2p1b2b/1PPpBB1q/P2Q1p1p/4PN2/1RK2P1P/8 w - - 0 1}{[Analyse on Lichess]}}

\textbf{Closest FENs - \href{https://lichess.org/analysis/1k1r4/1p2n3/1P1p4/3Pp2b/2Q1p3/2P5/3K1P2/q7 w - - 0 35}{[1]}}, \textbf{\href{https://lichess.org/analysis/1k1r4/p1p2p1p/6p1/P1B5/2N2n2/5B2/1PK2P1P/8 b - - 1 30}{[2]}}, \textbf{\href{https://lichess.org/analysis/3r3k/6p1/1p6/pPp1p1NP/2r5/4P3/1R3PK1/8 w - - 0 41}{[3]}}
\end{center}

\newpage
\adjustbox{max width=\appendixchessboardwidth\columnwidth}{
    \chessboard[
        showmover=true,
        color=yellow!70,
        setfen= r1b2rk1/p1q3pR/1pp1p3/1P4P1/Pn1p1PB1/b2Q1R2/1B1n2PP/6K1 w - - 0 1
    ]
    }
\begin{center}
    \textbf{\href{https://lichess.org/analysis/r1b2rk1/p1q3pR/1pp1p3/1P4P1/Pn1p1PB1/b2Q1R2/1B1n2PP/6K1 w - - 0 1}{[Analyse on Lichess]}}

\textbf{Closest FENs - \href{https://lichess.org/analysis/r4rk1/5ppp/1p6/2p1Pn2/p1PpNP1q/3Q1R2/1P4PP/3R2K1 w - - 0 28}{[1]}}, \textbf{\href{https://lichess.org/analysis/r1b2r1k/1p2q3/p3p1Q1/4P3/3PNn2/P4N2/1P4PP/R5K1 w - - 3 21}{[2]}}, \textbf{\href{https://lichess.org/analysis/r1b2rk1/1pR5/p3pp1p/2Qq2p1/3P1P2/P5R1/6P1/6K1 w - - 2 32}{[3]}}
\end{center}

\adjustbox{max width=\appendixchessboardwidth\columnwidth}{
    \chessboard[
        showmover=true,
        color=yellow!70,
        setfen= 2bB3k/1Pp5/Q1N2p2/P1r5/1q4p1/2bPRP1P/1r6/2R2K2 b - - 0 1
    ]
    }
\begin{center}
    \textbf{\href{https://lichess.org/analysis/2bB3k/1Pp5/Q1N2p2/P1r5/1q4p1/2bPRP1P/1r6/2R2K2 b - - 0 1}{[Analyse on Lichess]}}

\textbf{Closest FENs - \href{https://lichess.org/analysis/2B3k1/P4p1p/1N4p1/8/4p3/bp2P1P1/b4PP1/5K2 b - - 0 33}{[1]}}, \textbf{\href{https://lichess.org/analysis/6k1/5p1p/1R4p1/1P6/5p2/2b2P1P/1r4P1/2N2K2 b - - 4 43}{[2]}}, \textbf{\href{https://lichess.org/analysis/6k1/7p/p2N2p1/8/1q6/2bR3P/6P1/2R2K2 b - - 3 38}{[3]}}
\end{center}

\newpage
\adjustbox{max width=\appendixchessboardwidth\columnwidth}{
    \chessboard[
        showmover=true,
        color=yellow!70,
        setfen= r2q1rk1/3p1p2/pbn2p1B/n1q1PP2/2B1QR1p/1p4P1/P5KP/4R3 w - - 0 1
    ]
    }
\begin{center}
    \textbf{\href{https://lichess.org/analysis/r2q1rk1/3p1p2/pbn2p1B/n1q1PP2/2B1QR1p/1p4P1/P5KP/4R3 w - - 0 1}{[Analyse on Lichess]}}

\textbf{Closest FENs - \href{https://lichess.org/analysis/r2q1rk1/1R2p2p/2p3p1/2Pp4/3Q4/p5P1/P4n1P/4R1K1 w - - 0 26}{[1]}}, \textbf{\href{https://lichess.org/analysis/5rk1/5p2/p3p1pP/1pq1B3/5PR1/P1P4P/1P5K/8 w - - 0 36}{[2]}}, \textbf{\href{https://lichess.org/analysis/4r1k1/pp3p1p/2p2p2/4q3/2P1Q3/6P1/P4PKP/4R3 w - - 4 26}{[3]}}
\end{center}
\newpage
\onecolumn

%% file: neurips_2025_GDM_format.bbl
\begin{thebibliography}{3}
\providecommand{\natexlab}[1]{#1}
\providecommand{\url}[1]{\texttt{#1}}
\expandafter\ifx\csname urlstyle\endcsname\relax
  \providecommand{\doi}[1]{doi: #1}\else
  \providecommand{\doi}{doi: \begingroup \urlstyle{rm}\Url}\fi

\bibitem[Avni(1991)]{creativechess}
A.~Avni.
\newblock \emph{Creative Chess}.
\newblock Everyman Chess, 1991.

\bibitem[Levitt and Friedgood(1995)]{spectacularchess}
J.~Levitt and D.~Friedgood.
\newblock \emph{Secrets of Spectacular Chess}.
\newblock Everyman Chess, 1995.

\bibitem[Persson(2024)]{tigerchess}
T.~H. Persson.
\newblock \emph{Tiger's Chaos Theory}.
\newblock Quality Chess, 2024.

\end{thebibliography}
